\pdfoutput=1
\PassOptionsToPackage{backref=page,breaklinks=true}{hyperref}
\documentclass[11pt]{article}

\newif\ifauthordecided
\authordecidedtrue

\newif\ifarxiv
\arxivtrue

\newif\ifperfect
\perfectfalse

\ifarxiv
    \usepackage{acl}
\else 
    \usepackage[review]{acl}
\fi

\usepackage{times}
\usepackage{latexsym}

\usepackage[T1]{fontenc}

\usepackage[utf8]{inputenc}

\usepackage{microtype}

\usepackage{inconsolata}
\usepackage{graphicx}

\usepackage{booktabs}
\usepackage{pifont}

\usepackage{sec/package}

\definecolor{customblue}{HTML}{65aad3}
\definecolor{customlightblue}{HTML}{f0f6fc}
\definecolor{customblue2}{HTML}{2373b6}
\definecolor{customlightblue2}{HTML}{c3d9ee}

\newcommand{\tick}{\ding{51}}
\newcommand{\cross}{\ding{55}}

\newcommand{\ourdata}{\textit{Quriosity}\xspace}

\title{   
    \ourdata: Analyzing Human Questioning Behavior\\ and Causal Inquiry through Curiosity-Driven Queries
}

\ifauthordecided
\author{
Roberto Ceraolo\textsuperscript{1,}\thanks{\hspace{1.5pt} Equal contribution. Research done while Roberto was a research intern at ETH.}
\quad
Dmitrii Kharlapenko\textsuperscript{2,}\samethanks{}
\quad
Ahmad Khan\textsuperscript{2,}\samethanks{}
\\ 
\textbf{Amélie Reymond}\textsuperscript{\textbf{3}}
\quad
\textbf{Punya Syon Pandey}\textsuperscript{\textbf{4,5}}
\\ 
\textbf{
Rada Mihalcea\textsuperscript{6}
\quad
Bernhard Schölkopf\textsuperscript{7}
\quad
Mrinmaya Sachan\textsuperscript{2}
\quad
Zhijing Jin\textsuperscript{2,4,5,7}
}
\vspace{0.3em}
\\
\textsuperscript{1}EPFL
\quad 
\textsuperscript{2}ETH Zürich
\quad
\textsuperscript{3}University of Washington
\quad
\textsuperscript{4}University of Toronto
\quad
\\
\textsuperscript{5}Vector Institute
{ }
\textsuperscript{6}University of Michigan
{ }
\textsuperscript{7}MPI for Intelligent Systems, Tübingen, Germany 
\vspace{0.3em}
\\ 
\texttt{ceraolo.rc@gmail.com {} \{dkharlapenko,ahkhan,jinzhi\}@ethz.ch}
\\
}
\else
\author{Anonymous Authors}
\fi

\begin{document}

\maketitle
\begin{abstract}
Recent progress in Large Language Model (LLM) technology has changed our role in interacting with these models. Instead of primarily testing these models with questions we already know answers to, we are now using them for queries where the answers are unknown to us, driven by \textit{human curiosity}. This shift highlights the growing need to understand curiosity-driven human questions -- those that are more complex, open-ended, and reflective of real-world needs.
To this end, we present \ourdata, a collection of 13.5K naturally occurring questions from three diverse sources: human-to-search-engine queries, human-to-human interactions, and human-to-LLM conversations. Our comprehensive collection enables a rich understanding of human curiosity across various domains and contexts. Our analysis reveals a significant presence of causal questions (up to 42\%) in the dataset, for which we develop an iterative prompt improvement framework to identify all causal queries and examine their unique linguistic properties, cognitive complexity and source distribution. \ourdata paves the way for future work on causal question identification and open-ended chatbot interactions.
\footnote{Our code and data 
\ifarxiv
are at \url{https://github.com/roberto-ceraolo/quriosity}.
\fi
}
\end{abstract}

\section{Introduction}
\begin{center}
\textit{``Curiosity has its own reason for existing.''} \\
\end{center}
\makebox[\linewidth][r]{-- Albert Einstein}

Curiosity plays a crucial role in humans to motivate learning \cite{pmed-curiosity}. From a young age, children engage in curiosity-driven questioning to gain information about the world \cite{pmed-children-curiosity}. The rapid advancement of Large Language Models (LLMs) \cite{brown2020language,chowdhery2022palm} is transforming how we interact with them \citep{burns2024weak}. Rather than primarily acting as ``testers'' who pose questions to \textit{already} known answers, we are increasingly becoming ``inquirers'', asking questions that reflect \textit{genuine} curiosity. This transition, along with the widespread use of LLMs as chatbots and learning aids, necessitates a deeper understanding of how humans pose curiosity-driven questions \citep{ouyang2023shifted}.

From a computational social science perspective \cite{biester2024inferring,porter2016inferring}, it is intriguing to analyze the linguistic characteristics of these questions, the underlying human needs they express, and the user personality traits they reveal. Additionally, gaining better insight into the topics covered by these questions can aid in developing domain-specific methods to solve them.

However, most existing NLP datasets lack the comprehensive coverage of curiosity-driven questions from a diverse spectrum of contexts, as shown in \cref{tab:related_work}. These datasets often feature test questions such as 
``Who starred as Daniel Ruettiger in the film Rudy?'' \cite{rajpurkar2018know},
but numerous recent LLM queries explore complex topics, such as ``What are the causes of economic growth?'', which are absent in many existing datasets. While test questions are typically easier and designed to evaluate models, curiosity-driven inquiries are often more \textit{challenging}, \textit{open-ended}, and aligned with \textit{real-world needs}, driven by \textit{pure curiosity} \citep{coenen2019asking, rothe2018people, gottlieb2013information}. 

\begin{table*}[h]
    \centering
    \begin{adjustbox}{max width=\textwidth}
    \begin{tabular}{lcllcc}
        \toprule
        \textbf{Dataset} & \textbf{Curiosity Questions} & \textbf{Channels} & \textbf{Domain} \\
        \midrule
        SQuAD \citep{rajpurkar2018know} & \cross & Model Testing & Wikipedia \\
        ARC \citep{clark2018think} & \cross & Model Testing & Science \\
        TruthfulQA \citep{lin2022truthfulqa} & \cross & Model Testing & General Knowledge \\
        GSM8K \citep{cobbe2021training} & \cross & Model Testing & Math \\
        MMLU-Pro \citep{wang2024mmlu} & \cross & Model Testing & Complex Reasoning \\
        GPQA \citep{rein2023gpqa} & \cross & Model Testing & Expert Domains \\
        \midrule
        NaturalQuestions \citep{kwiatkowski2019natural} & \tick & H-to-SE (Search) & Web Search \\
        MSMarco \citep{bajaj2018ms} & \tick & H-to-SE & Web Search \\
        Quora Question Pairs \citep{quora2017} & \tick & H-to-H (Human-to-Human) & Q\&A Platforms \\
        ShareGPT\footnote{\url{https://huggingface.co/datasets/anon8231489123/ShareGPT_Vicuna_unfiltered}} & \tick & H-to-LLM & AI Interaction \\
        WildChat \citep{zhao2024inthewildchat} & \tick & H-to-LLM & AI Interaction \\
        \midrule
        \textit{\textbf{\ourdata (our work)}} & \tick & H-to-SE, H-to-H, H-to-LLM & A Variety of Domains \\
        \bottomrule
    \end{tabular}
    \end{adjustbox}
   \caption{Comparison of the source of questions for various datasets, including datasets commonly used in LLM testing. Given their curated nature, the questions in these datasets are not suitable for use in studying curiosity-driven questioning behavior.}
    \label{tab:related_work}
    \vspace{-1.0em}
\end{table*}

To bridge this gap, we present \ourdata, a comprehensive dataset of 13,500 naturally occurring questions derived from three diverse channels: human-to-search-engine queries (H-to-SE) from Google \cite{kwiatkowski2019natural} and Bing \cite{bajaj2018ms}, human-to-human interactions (H-to-H) from Quora \cite{quora2017}, and human-to-LLM interactions (H-to-LLM) from ShareGPT and WildChat \cite{zhao2024inthewildchat}. This multi-faceted approach ensures a rich tapestry of questions that reflect genuine human curiosity across various domains and contexts.

Our resulting dataset encompasses a wide variety of topics, and different cognitive complexities, from simple factual queries (e.g., ``How high is Mount Everest'') to complex causal investigations (e.g., ``What are the causes of economic growth?''). We find that natural inquiries are 38\% more open-ended than existing curated datasets, and also more equally cover a comprehensive range of six levels of cognitive complexity \cite{bloom1964taxonomy}.
Additionally, we find that people use various media to ask their questions based on different goals. Specifically, H-to-SE queries are typically motivated by knowledge and information needs, whereas H-to-LLM questions more frequently address problem-solving and leisure-oriented purposes.

Furthermore, we identify an important phenomenon, emergent causal inquiries \cite{pearl2009causal,peters2017elements}, within natural questions.
Our data analysis reveals that up to 42\% of the questions are related to causality. These questions are particularly significant since curiosity-driven questions are often asked with a future action in mind. These types of questions are inherently causal, thus making causal questions especially meaningful for humans \citep{pearl2019seven, sloman2015causality}.

To systematically study these queries, we develop an iterative prompt improvement method for identifying and categorizing causal questions.
Based on this categorization, we analyze the distinct linguistic properties, cognitive complexity, and source distributions of causal questions. Finally, we lay the groundwork for future research by investigating the performance of current LLMs on our question set and building efficient causal question classifiers as routers to differentiate between causal and non-causal questions, directing them to an enhanced reasoning pipeline, as the literature demonstrates the efficiency of routing techniques to enhance performance \citep{chen2023frugalgpt}.

\vspace{-0.005em}
In summary, our contributions are as follows:
\begin{enumerate}
    \item We present \ourdata, a new collection of 13,500 curiosity-driven human queries across three diverse sources.
    \item We identify the differences between curiosity-driven questions in our \ourdata dataset and existing curated NLP datasets in terms of open-endedness, cognitive complexities,  user needs, and knowledge domains.
    \item 
    We further explore an important phenomenon in curiosity-driven questions---causal inquiries---and analyze its distinct linguistic properties, cognitive complexity, and distribution across different sources.
    \item 
    To identify causal questions across \ourdata, we present an iterative prompt improvement framework combined with a limited set of human expert labels to scale up causal question labeling in our large dataset.
    \item 
    We provide preliminary studies on efficient causal question classification using six non-neural-network or small language models and LLM response analysis.
\end{enumerate}

\section{Dataset Construction}
\subsection{Question Sources}\label{sec:collection_process}

To compose our \ourdata dataset, we identify three different channels where people ask questions online: people enter queries on search engines (H-to-SE), post questions to other people on question platforms (H-to-H), and chat with LLMs hosted on web interfaces such as ChatGPT (H-to-LLM). To represent H-to-SE data, we obtain search queries on Google from the NaturalQuestions dataset \citep{kwiatkowski2019natural}, and on Bing Search from the MSMarco dataset \citep{bajaj2018ms}. For the H-to-H data, we adopt the Quora Question Pairs dataset \cite{quora2017},\footnote{\url{https://quoradata.quora.com/First-Quora-Dataset-Release-Question-Pairs} } which compiles actual question data from the widely-used question-answering website Quora. Lastly, we also incorporate several sources of H-to-LLM queries, from ShareGPT, a collection of users' voluntarily-shared queries to ChatGPT, and the WildChat collection of user-LLM conversations through their chat interface powered by ChatGPT and GPT-4 APIs \cite{zhao2024inthewildchat}. Questions asked on these channels are typically curiosity-driven; while some H-to-SE or H-to-LLM questions might be asked to test the capabilities of these services, this is likely to only represent a small portion of these queries.

\begin{table}[ht]
\centering \small
\setlength\tabcolsep{3.5pt}
\begin{tabular}{llccc}
\toprule
\textbf{Nature of Data} & \textbf{Dataset} & \textbf{{\# Samples}}
\\
\midrule %
\multirow{2}{*}{H-to-SE (33.3\%) } & MSMarco (\citeyear{bajaj2018ms}) & 2,250 
\\
& NaturalQuestions (\citeyear{kwiatkowski2019natural}) & 2,250         
\\
\midrule %
H-to-H (33.3\%)  & Quora Question Pairs & 4,500       
\\
\midrule %
\multirow{2}{*}{H-to-LLM (33.3\%) } & ShareGPT  & 2,250   
\\
& WildChat (\citeyear{zhao2024inthewildchat})        & 2,250       
\\
\bottomrule
\end{tabular}
\caption{Our \ourdata equally covers questions from the three source types: human-to-search-engine queries (H-to-SE),
human-to-human interactions (H-to-H), and human-to-LLM interactions (H-to-LLM).}
\label{tab:sources}
\vspace{-0.5em}
\end{table}

To preprocess the data, we filter out empty queries, non-English questions, and invalid characters. For LLM conversations, we select the first question and cut off follow-up questions. For long questions with more than 30 words, we generate shorter versions using GPT3.5-turbo-0125 to condense the main idea of the questions. As shown in \cref{tab:sources}, we sample 4,500 questions from each of the three channels, and if a channel has multiple sources, we also distribute the questions evenly across the sources. This results in a total of 13,500 questions for our \ourdata dataset, which can be accessed at \ifarxiv \url{https://huggingface.co/datasets/causal-nlp/Quriosity}\fi.

\subsection{Data Statistics}

\paragraph{Overall.} 
Our dataset has a total of 13,500 samples, each with an average length of 11.34 worde. The entire vocabulary size of our data is 25.7K unique words, with a type-token ratio (TTR) of 0.168. We describe the statistics of our dataset in the appendix in \cref{tab:stats_all}. 

\begin{figure}[tbp]

    \centering
    \includegraphics[width=\linewidth]{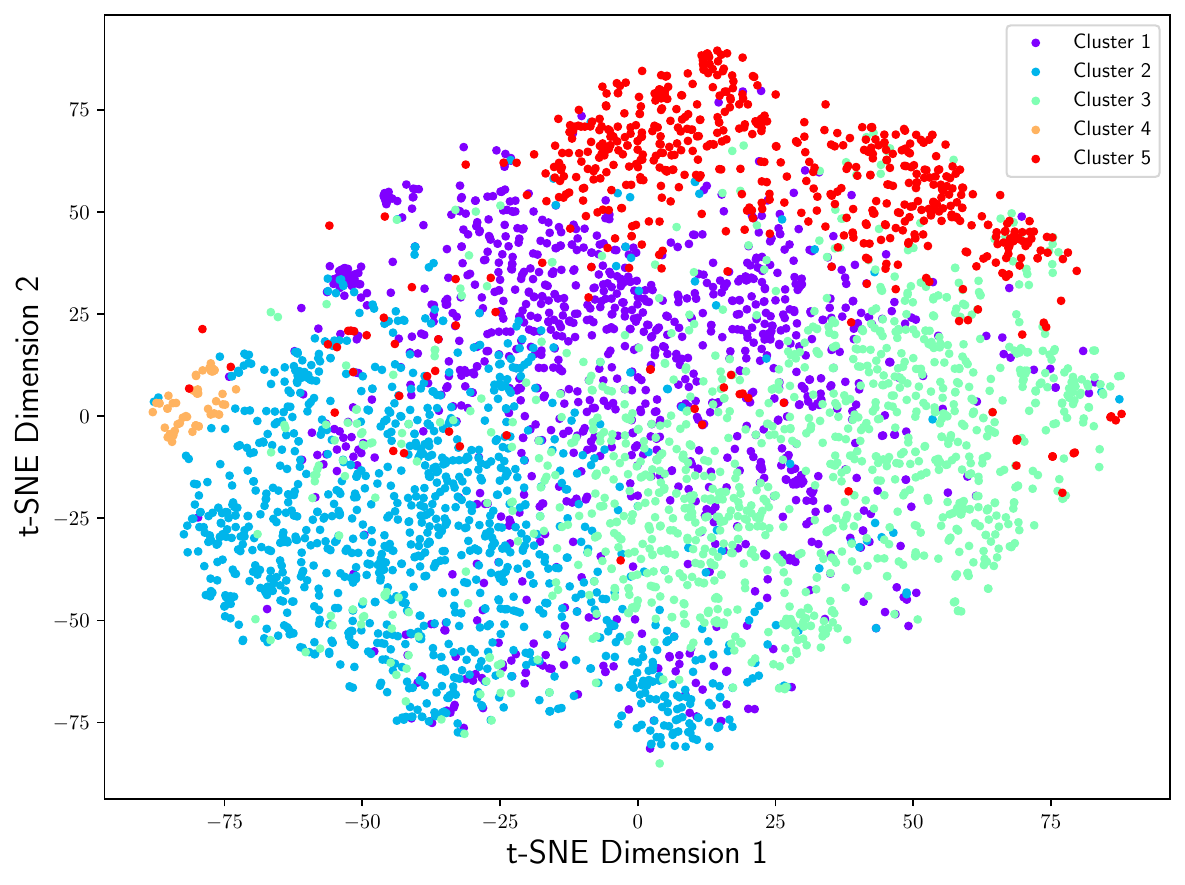}
    \caption{T-SNE visualisation of the main topic clusters in our \ourdata dataset. Cluster 1: Daily life. Cluster 2: Computer-related. Cluster 3: Sports, medicine, and science. Cluster 4: Prompt Questions. Cluster 5: Stories and fictional characters. See examples for each cluster in \cref{tab:cluster-examples}.}
        \label{fig:embeddings} 
\vspace{-0.5em}
\end{figure}

\paragraph{Topic Coverage.} 
To explore the topics covered in our dataset, we perform K-means clustering \citep{hartigan1979algorithm} on the embedding space of all queries using the text embedding model \textit{text-embedding-3-small} from OpenAI. \cref{fig:embeddings} uses t-SNE \citep{van2008visualizing} to visualize the five main clusters: daily life questions, computer-related inquiries, sports, medicine and science, prompt questions, and story generation.

\subsection{Comparison to Existing Datasets}
In constructing \ourdata, we wanted to capture a broad spectrum of curiosity-driven questions. As contrasted in \cref{tab:related_work}, existing datasets are predominantly made up of questions that serve a testing purpose. Here, we make comparisons to popular benchmark datasets.

Popular datasets cover a broad range of tasks and topics: SQuAD covers reading comprehension tasks \citep{rajpurkar2018know}, ARC covers multiple-choice science questions \citep{clark2018think}, TruthfulQA covers truthful question answering \citep{lin2022truthfulqa}, GSM8K consists of grade school math word problems \citep{cobbe2021training}, MMLU-Pro questions require complex reasoning \citep{wang2024mmlu} and GPQA questions require domain experts \citep{rein2023gpqa}.

Despite this coverage, all of these datasets share similarities such that questions are designed to test LLMs. These questions are not driven by genuine human curiosity, and thus cannot help us understand human questioning patterns better. We note that these datasets could include questions that can be asked by a curious human, e.g. in TruthfulQA or MMLU-PRO. However, due to the curated design of these datasets, they do not work as a good representative sample, and analysis around curiosity-driven questions cannot be made from these datasets.

In contrast, the questions we used to develop \ourdata come from the NaturalQuestions, MSMarco, Quora Question Pairs, ShareGPT and WildChat datasets. The questions in each dataset meet our criteria for being curiosity-driven, as they were written naturally by humans seeking knowledge rather than designed for model evaluation. However, these datasets are not representative of all the channels of human queries (i.e. H-to-SE, H-to-H, H-to-LLM) and are thus not individually representative. Additionally, some of the datasets had to be repurposed to extract curiosity-driven questions (e.g. for Quora Question Pairs, ShareGPT, and Wildchats). Thus overall, our dataset provides the first collection of a representative sample of curiosity-driven questions by humans.

\section{Exploring Curiosity-Driven Queries}

\subsection{How Do Natural Inquiries Differ from Curated Test Questions?}
\label{sec:natural_vs_tests}
Our dataset enables a rich set of explorations on curiosity-driven questioning behavior.
Specifically, we explore two research questions:  what are the properties distinguishing curiosity-driven inquiries from curated test questions (in this section), and how does human behavior vary across different platforms (in \cref{sec:platform})?

To compare curiosity-driven questions in \ourdata with curated ones, we collect a comparison set consisting of six curated test sets. We use the same datasets mentioned in \cref{tab:related_work} that were not used in constructing \ourdata, namely SQuAD, ARC, TruthfulQA, GSM8K, MMLU-Pro and GPQA. We sample 500 questions from each of them, to build a representative sample of popular curated questions. 

\paragraph{Methods.}
We compare curiosity-driven and curated test queries in terms of cognitive complexity and open-endedness. For cognitive complexity, we adopt the six levels of cognitive abilities in Bloom's Taxonomy
\cite{anderson2001taxonomy}: starting with the simplest skill, remembering, and advancing to understanding, applying, and creating. For open-endedness, we evaluate whether the question permits semantically different answers (i.e., open-ended) or only has one unique answer (i.e., not open-ended).
Both evaluations are implemented using LLMs (GPT-4o-mini with prompts shared in \cref{app:prompts}), within an LLM-as-a-judge framework \cite{zheng2023judging}. We further verify the quality of the LLM annotations by calculating its agreement with human annotation on a small set of 150 samples, where we obtain a reasonable F1 score of 79\%.

\begin{table}[th]
\centering \small
\begin{tabular}{lcc}
\toprule
\textbf{Category} & \textbf{Curated Questions} & \textbf{\ourdata} \\
\midrule
\multicolumn{3}{l}{\textit{\textbf{Open-Endedness}}}\\
Open-Ended & 30\% & {68\%} \\
\hline
\multicolumn{3}{l}{\textit{\textbf{Cognitive Complexity}}}\\
Remembering     & 30.51\% & 36.82\% \\
Understanding   & 7.47\%  & 13.47\% \\
Applying        & 36.41\% & 13.54\% \\
Analyzing       & 13.90\% & 8.8\%  \\
Evaluating      & 11.54\% & {13.74\%} \\
Creating       & 0.17\%  & {13.62\%} \\
\bottomrule
\end{tabular}
\caption{Comparison of curated test questions and our \ourdata in terms of open-endedness and cognitive complexity. Remembering is the least complex, and Creating is the most complex.}
\label{tab:non-nat-vs-ours_quriosity}
\vspace{-0.5em}
\end{table}

\paragraph{Results.}
Comparing the curated questions and our \ourdata in \cref{tab:non-nat-vs-ours_quriosity}, we find that curiosity-driven questions tend to be much more open-ended, with 68\% questions allowing semantically different answers. Moreover, curiosity-driven questions often require a more even distribution of all the six levels of cognitive skills, with each skill's portion closer to 13\%, and a higher portion of advanced skills such as evaluating and creating.

\subsection{Does Human Question Behavior Vary across Channels?}\label{sec:platform}

We are further interested in inferring human behavior from the questions, and how it varies across the three channels of inquiries: H-to-SE, H-to-H, and H-to-LLMs. We start our analysis by inferring user needs and intent from the questions, and compare them across the three channels. We then analyze the differences in cognitive complexity and knowledge domains of the questions across the channels.

\subsubsection{Inferred User Needs}\label{sec:need}
\begin{figure}[h]
    \centering
    \includegraphics[width=\linewidth]{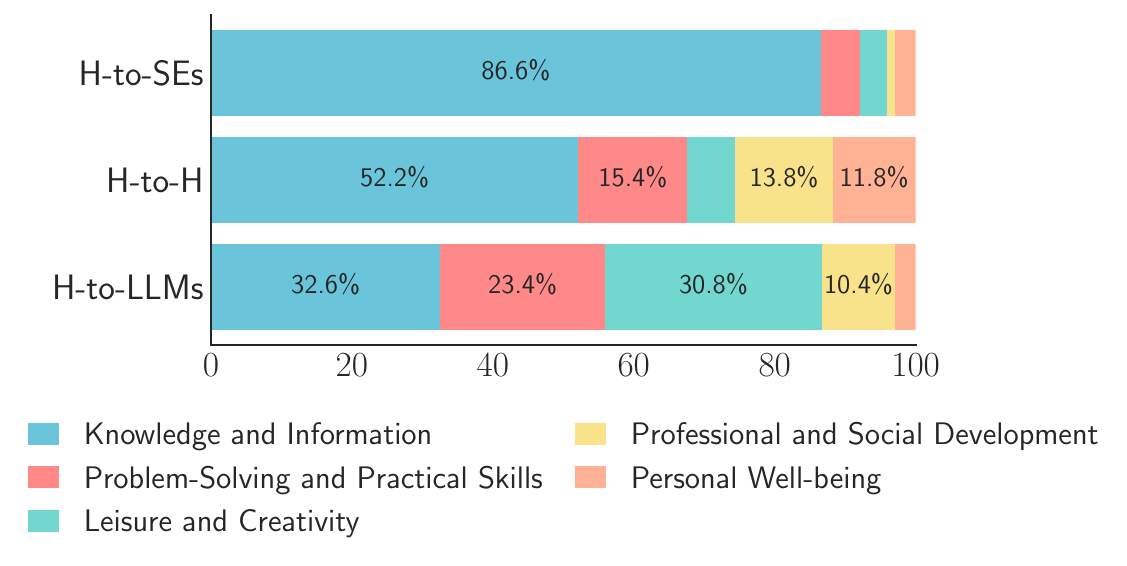}
    \label{fig:need}
    \caption{User needs across sources of H-to-SE, H-to-H, and H-to-LLM interactions.}
\end{figure}

Inspired by existing frameworks of human needs \citep{maslow1943theory} and information seeking behavior \citep{wilson1999models, kuhlthau2005information}, we introduce five types of needs relevant for \ourdata: ``Knowledge and Information'', ``Problem-Solving and Practical Skills'', ``Personal Well-being'', ``Professional and Social Development'', and ``Leisure and Creativity''. To label the needs following our taxonomy, we annotate a small set, and then use LLMs to iteratively improve the prompt, before scaling up LLM labels to the entire set, with details in \cref{app:prompts}. 

We plot in \cref{fig:need} the distribution of user needs inferred from questions across the three channels. Notably, we find that the basic need of ``Knowledge and Information'' is dominant in H-to-SE queries, but much less in others, e.g., as low as 32.6\% in H-to-LLMs interactions. 
In contrast, LLMs are used more often to address the needs of ``Leisure and Creativity'' (30.8\%) and ``Problem-Solving and Practical Skills''  (23.4\%). These results show a shift in user expectations from AI systems in contrast to search engines from simple factual queries to more creative and interactive problem-solving. 
Future work can conduct additional longitudinal studies to investigate whether the advancement of LLMs has a causal effect to change user's choice of the medium to ask questions.

\subsubsection{Cognitive Complexity} \label{sec:bloom_across_platforms}

\begin{table}[h]
\centering \small
\setlength\tabcolsep{2pt}
\begin{tabular}{p{3.2cm}ccc}
\toprule
\textbf{Category} & \textbf{H-to-SE} & \textbf{H-to-H} & \textbf{H-to-LLMs} \\
\midrule
\multicolumn{4}{l}{\textit{\textbf{Cognitive Complexity}}}\\
Remembering     & \textbf{76.49\%} & 19.42\% & 14.52\% \\
Understanding   & 13.82\% & 15.27\% & 11.32\% \\
Applying        & 4.07\%  & 18.73\% & 17.81\% \\
Analyzing       & 3.40\%  & 13.00\% & 10.05\% \\
Evaluating      & 1.87\%  & \textbf{31.27\%} & 8.09\%  \\
Creating        & 0.36\%  & 2.31\%  & \textbf{38.20\%} \\
\hline
\multicolumn{4}{l}{\textit{\textbf{Knowledge Domain}}}\\
Arts and Culture                       & 13.89\% & 5.30\%  & 12.54\% \\
Computer Science                       & 6.80\%  & 17.87\% & \textbf{45.11\%} \\
Everyday Life \& Personal Choices     & 9.63\%  & 21.57\% & 11.91\% \\
Health and Medicine                    & \textbf{27.99\%} & 10.94\% & 2.71\%  \\
Historical Events \& Hypothetical Scenarios & 10.04\% & 3.22\%  & 3.44\%  \\
Natural \& Formal Sciences            & 13.39\% & 7.91\%  & 4.84\%  \\
Psychology \& Behavior                & 0.71\%  & 11.47\% & 3.82\%  \\
Society, Economy \& Business             & 17.55\% & \textbf{21.73\%} & 15.63\% \\
\bottomrule
\end{tabular}
\caption{Distribution of cognitive complexity and domain classifications for different sources (H-to-SEs, H-to-H, H-to-LLMs).}
\label{tab:channel_distribution_quriosity}
\end{table}

We show the three channels' distribution across the six levels of cognitive complexity in Bloom's Taxonomy
\cite{anderson2001taxonomy} in \cref{tab:channel_distribution_quriosity}.
We find that in H-to-SEs, most questions (76.49\%) fall under the Remembering category, indicating that users primarily seek factual information retrieval from search engines. This resonates with the dominant ``Knowledge and Information'' need in \cref{sec:need}. In contrast, among H-to-H interactions, the majority of questions are ``Evaluating,'' -- requiring subjective judgments and nuanced understanding, which explains our dataset's open-endedness illustrated earlier in \cref{sec:natural_vs_tests}. In H-to-LLMs, the largest category is Creating (38.20\%), suggesting that users frequently request generative and novel content from LLMs. 

\subsubsection{Knowledge Domain}

We use LLMs to identify the main knowledge categories in \ourdata in \cref{tab:channel_distribution_quriosity}. The result suggests that humans predominantly use LLMs for computer science-related questions (45.11\%), while search engines receive more inquiries related to health and medicine (27.99\%). On the other hand, humans ask each other a more balanced variety of questions, with ``Society, Economy, Business'' (21.73\%) and ``Everyday Life and Personal Choices'' (21.57\%) being the most frequent. This indicates that users (currently) rely on LLMs for technical queries, on search engines for health-related information, and on human interactions for broader, context-rich discussions.

\section{Identifying Causal Inquiries}
In this section, we will introduce how we identify causal inquiries among all the curiosity-driven questions, their distinct features, and correlations with the other behaviors.
\subsection{Why Causality?}
As motivated in the introduction, an interesting phenomenon in human queries is their causality-seeking behavior. Rooted in philosophy \cite{beebee2009oxford,russell2004history,kant1781critique}, causality has evolved into a rigorous statistical field \cite{fisher1927,rubin1980,spirtes1993causation,pearl2009causality}. Causal questions are a particularly significant category of curiosity-driven questions since such questions often involve an anticipatory action, extending toward generating new information for unknown prompts. Since LLMs have been shown to not answer causal questions very well \cite{jin2023cladder},studying these questions is important to help us design LLMs that answer these questions better, with findings extending across naturally occurring questions.

\subsection{Causal Query  Identification}

\paragraph{Human Annotation of Causal Questions}
Given the formal formulation of causality, we engaged two expert annotators with extensive knowledge in both causality and NLP to annotate a sample of  $N=500$ data points. 
They were briefed on the nature of the data and the potential presence of sensitive content. Additionally, the annotators represent diverse genders and cultural backgrounds. The annotator instructions can be found in \cref{app:annotator_instructions}.

We develop the following iterative improvement process: Using the provided annotation guideline, annotators independently label 500 data points. Then, we check the inter-annotator agreement rate, achieving a Cohen's $\kappa$ \cite{cohen1960coefficient} of 0.66, indicating moderate consensus among the human labelers \cite{mchugh2012interrater, landis1977application, viera2005understanding}.
The annotators then further refine the labels by analyzing the disagreement cases and either: (1) In cases where the initial textual guideline does not communicate the mathematical formulation clearly, we improve the guidelines, and the annotators agree on a clear classification choice. (2) In cases where the two annotators interpret the data sample differently, they have a discussion and agree on one correct label for each question. The agreed-upon label then becomes the ground truth. We identify 238 out of 500 annotated samples as causal questions.

\paragraph{Scaled Labeling by Iterative Prompt Improvement}
\label{sec:prompt_improvement}

Using the human annotation set, we develop an iterative prompt improvement process on the entire dataset. We start the process by first providing our definitions of causal questions. After this initial annotation by GPT-4 on the sample set, we calculate the classification performance of the LLM with regard to our ground-truth labels, and inspect the error cases manually. Then, we improve the prompt by the canonical prompt engineering techniques including in-context learning \citep{brown2020language} and chain-of-thought prompting \citep{wei2022chain}), and by clarifying the decision principles around the error cases. We repeat this iterative improvement process until we reach a high classification performance.\footnote{We report in \cref{sec:prompt_iteration_examples} some examples of early-iteration vs. late-iteration prompts.} The final performance of GPT-4 reaches 88.7\% F1 scores with regard to the ground-truth causal question labels in the 500 sample set, 89.4\% accuracy, 92.9\% precision, and 84.9\% recall. 
On a held-out set of 100 additional annotated samples not belonging to the previous set, we observe a good performance of 88\% F1 score.

\subsection{How Do Causal Questions Differ from Non-Causal Ones?} \label{sec:causal_vs_noncausal}
\label{sec:data_analysis}

Based on our scaled labels of causal questions, we investigate two research questions: how  causal  questions differ from non-causal questions (in this section), and how causal questions are distributed (in \cref{sec:causal_distr}).

\subsubsection{Linguistic Differences}
\paragraph{Method.} We first explore the differences between causal and non-causal queries by  linguistic features. First, we calculate the frequency of different question words in causal and non-causal questions. We further follow \citet{girju2002mining} to identify the presence of morphological causatives (verbs ending in \textit{-en} or \textit{-ify}), lexico syntactic patterns (presence of causative verb phrases such as \textit{lead to, induce, result in}), and matching rules by \citet{Bondarenko2022CausalQA}. 

\begin{table}[h]
\centering \small
\begin{tabular}{lcccc}
\toprule
{Question Words} &  {Causal} & {Non-Causal} & {Overall} \\
\midrule
How & \textbf{2,077} & 636 & 2,713 \\
Why & \textbf{635} & 12 & 647 \\
\hline
What & 1,383 & \textbf{2,324} & 3,707 \\
Who & 144 & \textbf{723} & 867 \\
Where & 124 & \textbf{465} & 589 \\
\bottomrule
\end{tabular}
\caption{Distribution of question words in our dataset.}
\label{tab:linguistic}
\vspace{-0.0em}
\end{table}

\begin{table}[h]
\centering
\small
\begin{tabular}{lccc}
\toprule & Causal & Non-Causal & Overall \\
\midrule
Morphological & \textbf{1.65\%} & 0.56\% & 1.02\% \\
Lexico-Syntactic & \textbf{11.52\%} & 6.83\% & 8.81\% \\
Causative Rules & \textbf{15.21\%} & 0.40\% & 6.65\% \\
\bottomrule
\end{tabular}
\label{tab:causal_indicators}
\caption{Distribution of causal indicators in questions, including morphological causative,  lexico-syntactic pattern, and causative rules.}
\vspace{-1.5em}
\end{table}

\paragraph{Results.} 
As shown in \cref{tab:linguistic}, causal questions are more frequently led by question words such as ``Why'' and ``How,'' whereas non-causal questions often start with ``What,'' ``Who'' and ``Where.'' 
In terms of other linguistic indicators in \cref{tab:causal_indicators}, causal questions have more morphological causatives and lexico-syntactic patterns, as well as matches with more causative rules. However, we note that the three linguistic methods capture a limited set of causal questions, which indicate the linguistic diversity and the richness of our \ourdata dataset.

\subsubsection{Cognitive Complexity}
\label{sec:causal-vs-nc-cognitive-comp}
\begin{figure}[h]
    \centering
    \includegraphics[width=\linewidth]{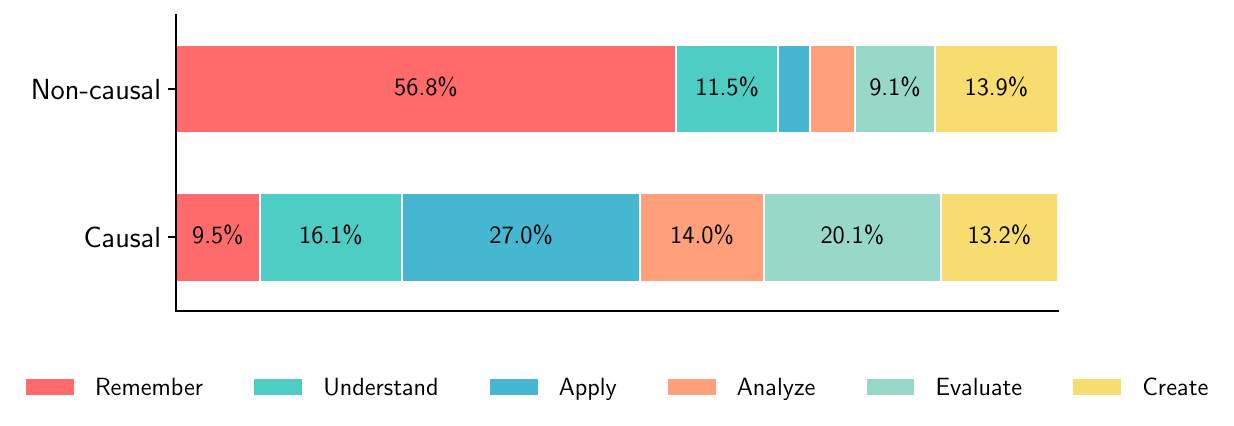}
    \vspace{-1em}
    \caption{Cognitive skill distribution in causal vs. non-causal questions, showing a more balanced distribution across causal inquiries.}
    \label{fig:bloom}
\end{figure}

Another distinction of causal questions is that it requires higher-level cognitive capabilities. We present the cognitive skill distribution across causal vs. non-causal questions in \cref{fig:bloom}, which shows that non-causal questions rely more on lower cognitive requirements like Remembering, whereas causal questions tend to require higher cognitive skills like Applying, Analyzing, and Evaluating.

\subsection{How Are Causal Questions Distributed?}\label{sec:causal_distr}
Connecting causal queries with our previous findings, we analyze how causal questions are distributed across curiosity-driven vs. curated questions, and across the three different platforms.

\paragraph{Causality in Curiosity-driven vs. Curated Questions}
Comparing the percentage of causal questions across curated test sets and our \ourdata, we find that our curiosity-driven questions almost double the percentage of causal queries (42\%) than that of curated questions (23\%) in \cref{tab:distr}.
This again highlights a contrast between curiosity-driven queries made by humans and crafted tests. Our findings reinforce previous work's insight on the divergence between user queries and NLP benchmarks \cite{ouyang2023shifted}.

\begin{table}[h]
\centering \small
\begin{tabular}{lcc}
\toprule
 & {Causal} & {Non-Causal} \\
\midrule
{Curated} & 23\% & 77\% \\
{\ourdata} & 42\% & 58\% \\ \hline
{H-to-SE} & 22\% & 78\% \\
{H-to-H} & 59\% & 41\% \\
{H-to-LLMs} & 46\% & 54\% \\
\bottomrule
\end{tabular}
\caption{Distribution of causal questions across natural and non-natural questions, as well as the three platforms.} 
\label{tab:distr}
\end{table}

\paragraph{Causal Queries across Platforms}
Identifying causality in questions across the three platforms in \cref{tab:distr}, we find that H-to-H has the largest proportion of causal questions at 59\%, followed by H-to-LLMs at 46\%, and finally, H-to-SE at 22\%. This pattern suggests that when seeking to understand causes and effects, internet users view LLMs as more aligned with human responses compared to search engines. In the future, there might be an increasing transition towards conversational systems for information retrieval and understanding of the world \citep{zhou2024understanding}.

\section{Paving the Way for Future Work}

\subsection{Building Efficient Causal Question Routers}

As shown earlier in the study, causal questions have their own unique nature, and occupy a non-trivial percentage, 42\%, of curiosity-driven questions. Moreover, as we show in the next section, LLM performance on answering causal questions is quite limited. Given this motivation, we imagine an emerging future direction is to classify causal questions beforehand, and potentially route them to some specific, reasoning-enhanced solution pipeline.

To efficiently identify causal questions, we explore a set of seven smaller, more efficient models in \cref{tab:training_results}, trained or fine-tuned on \ourdata to classify causal questions from non-causal ones, aiming to understand this task's tradeoff between model size and accuracy. Such classification models have proven effective for router construction. \citep{ong2024routellm, ding2024hybrid}.

We see that the largest model, FLAN-T5-XL (LoRA), performs best. However, a much smaller model, FLAN-T5-Small, also provides a good compromise with a small drop in accuracy but significant computation savings. Details on our experiments and further analysis can be found in \cref{app:causal-classifiers}. We also include a comparison to Deepseek R1 7B model \cite{deepseekR1} as a baseline. This work provides a starting point for future work to classify causal questions, and potentially build dedicated causal reasoning modules.

\begin{table}[ht]
    \centering \small
    \setlength\tabcolsep{2pt}
    \begin{tabular}{lcccccccc}
    \toprule
         & \multirow{1}{*}{\# Params} & 
         \multirow{1}{*}{F1} & \multirow{1}{*}{Acc.} & \multirow{1}{*}{P.} & \multirow{1}{*}{R.} \\
    \midrule
    Rule-based classifier & --
    & 58.7 & 23.3 & 62.8 & 55.2 \\
    Deepseek R1 & 7B  & 64.2 & 74.1 & 77.1 & 55.0 \\
    TF-IDF + XGBoost & --
    & 72.3 & 79.0 & 78.1 & 67.4 \\
    FLAN-T5-Small & 80M
    & 84.2 & 86.5 & 83.3 & 85.1 \\
    FLAN-T5-Base  & 250M
    & 85.7 & 87.9 & 85.5 & 86.0 \\
    FLAN-T5-Large (LoRA) & 780M
    & 85.8 & 88.0 & 85.0 & 87.0 \\
    Phi-1.5 (LoRA) & 1.5B
    & 85.3 & 87.3 & 83.4 & \textbf{87.2} \\
    FLAN-T5-XL (LoRA) & 2.85B
    & \textbf{87.7} & \textbf{89.2} & \textbf{91.1} & 85.4 \\
    \bottomrule
    \end{tabular}
    \label{tab:training_results}
    \caption{Performance of efficient models
    on causal question identification, with Deepseek R1 as a baseline.
    We report each model's computational cost, accuracy (``Acc.''), and F1,  precision (``P.'') and recall (``R.'') scores w.r.t. the causal class. We put ``--'' for the non-LLM models for its low computational cost.
    }
    
\end{table}

\subsection{Suggesting Improvement Directions for Future LLMs}
\label{sec:future-dir1}
While our study presents an extensive investigation of human questioning behavior, a natural next step is to evaluate the capability of LLMs to answer these curiosity-driven questions.
To this end, we provide some preliminary work in evaluating the performance of the latest LLM, GPT-4o, based on three usability criteria commonly used in user satisfaction surveys \citep{international2018ergonomics}: effectiveness, efficiency, and satisfaction. On a scale of 1--5, we annotate the answer quality of 50 random causal questions, and find that
GPT-4o obtains an average effectiveness of 3.83, efficiency of 2.88, and satisfaction of 3.85. We find it struggles to foresee anticipatory causal effects, but does well in answering causal questions that require knowledge lookup. Its answers are often overly verbose, explaining its limited score in efficiency (\cref{app:answer-eval-gpt4o}).

\section{Related Work}
\paragraph{Socio-Linguistic Analysis of Human Behavior Analysis}
Psychologists and linguists have long been interested in the study of questions, as they provide valuable insights into human cognition, emotion, and social dynamics. Traditionally, this task has been approached through methods such as manual content analysis \citep{graesser1994question}, discourse analysis \citep{sinclair2013towards}, and corpus linguistics \citep{biber2012register}. More recently, there has been a shift towards computational methods and large-scale data analysis, as evidenced by the development of tools like Linguistic Inquiry and Word Count (LIWC) \citep{tausczik2010psychological} and the application of natural language processing techniques \citep{boyd2021natural}. \citet{pennebaker2003psychological} examine how natural language use, including questioning behavior, reflects personality traits and social processes. \citet{jackson2022text} explore how analyzing language, including questions, can advance psychological science by providing insights into cognitive and emotional processes. To our knowledge, existing research falls short in inferring insights about human curiosity in information-seeking behavior.

\paragraph{Causal Question Datasets}

There is a growing research interest in causal reasoning of LLMs \citep{zhang2023understanding,kıcıman2023causal, zečević2023parrots,jin2023cladder,jin2024large}. However, existing literature focuses on ``test'' questions, lacking coverage of curiosity-driven ``inquiries'' and a comprehensive collection of curiosity-driven causal questions. Indeed, while some previous studies focus on a subset of causality \citep{jin2023cladder, tandon2019wiqa, gusev2022headline}, others only use linguistic heuristics to label a question as ``causal'' \citep{lal2021tellmewhy,verberne2006discourse, verberne2008using, zong2021tell}, primarily including artificially-generated data \citep{Bondarenko2022CausalQA, mostafazadeh2020glucose, roemmele2011choice}, rarely including sources of curiosity-driven questions. Moreover, since most datasets were collected before the rise of LLMs, none of them include new sources of curiosity-driven questions -- causal questions directly asked to LLMs \citep{ouyang2023shifted}.

\section{Conclusion}
This study identified the gap between previously curated NLP datasets and curiosity-driven human questions. With a collection of 13.5K questions in \ourdata, we analyzed the distinct features of curiosity-driven questions and further explore the properties of causal inquiries. Our work presents an up-to-date question set in the era of LLMs, and paves the way for future model improvements on causal reasoning.

\section*{Limitations}
\label{sec:limitations}
\paragraph{Selection bias}
\label{sec:selection_bias}

\ourdata is the first effort towards building a representative sample of causal questions humans ask online, to study human curiosity. Clearly, the data selection process is not devoid of bias. By \textit{bias} here we imply that there might be a difference between the set of questions we gathered (\ourdata) and the full set we are trying to make an inference about (human curiosity as a whole) \citep{lin2020language}. Before analyzing the bias for the full set of questions humans ask, we wonder about a subset: how representative is \ourdata of the questions internet users ask? Each of the sources we used was gathered by researchers independently, so they underwent different filtering procedures, which might affect the distribution of questions. The following are potential sources of bias for each of the components of \ourdata: 
\begin{enumerate}
\item ShareGPT: questions that yielded a denial from ChatGPT were excluded (e.g. ``\textit{I'm sorry but}'')
\item NaturalQuestions: among Google queries, a subset of questions was selected with specific syntactical patterns, yielding a Wikipedia page among the top 5 Google results. 
\item MSMARCO: only questions on Bing looking for a specific answer, and for which human judges could generate an answer based on some retrieved text passages were included. 
\item WildChat: since the conversations were collected on Huggingface Spaces, most likely the average user was a developer, or someone involved in the AI community - and not the general internet population at large
\item Quora Question Pairs: the authors declare they used sanitation methods such as removal of questions with long question details
\end{enumerate}

The above filtering procedures constitute a limitation of this work. In the future, the inclusion of more varied data sources can help reduce the bias coming from any single source. 

Ideally, we would like to generalize such insights to humans, not just internet users. Can we consider the questions asked online as a representative sample of the full set of human questions? If an individual has a question, there might be several reasons why she does not ask it online and hence does not leave a trace. Those reasons act as confounders and generate a bias in the questions found online. For instance, users concerned about privacy might decide to avoid asking private questions on search engines or to LLMs. Future work could include more data sources or involve conducting surveys that aim to identify and categorize questions not typically found online, thus potentially decreasing the influence of these biases. Also, having filtered out all non-English queries, the insights mainly apply to the countries where people tend to use the Internet in English. Future work could focus on building a multilingual-\ourdata to improve its coverage. 

\paragraph{Limitations of the Classifiers}
While our causal question classifiers show promise, they have several limitations. As \ourdata is the first dataset with causal question labels, external validation was not possible, highlighting the novelty of our work but also the need for additional labeled datasets in this domain. We focused primarily on binary classification without analyzing performance across different causal question subcategories. Additionally, we did not explore the impact of this classification on downstream tasks such as question answering or causal inference. Although we compared our models with a rule-based classifier, more comprehensive comparisons with other methods adapted for causal question identification could provide further insights. Finally, \ourdata may contain inherent biases due to its online sources, potentially influencing classifier performance and generalizability. Addressing these limitations in future work could involve creating additional labeled datasets, developing fine-grained classification models, exploring practical applications, and establishing standardized benchmarks for causal question identification.

\section*{Ethical Considerations}
\paragraph{Data License}\label{sec:ethical}
\ourdata comprises publicly available sources. We carefully review the licenses of each source used to build \ourdata. The \href{https://huggingface.co/datasets/anon8231489123/ShareGPT_Vicuna_unfiltered}{shareGPT} dataset has a \textit{Apache-2} license, WildChat has a \href{https://allenai.org/licenses/impact-lr}{\textit{AI2 ImpACT License – Low Risk Artifacts}}, MSMARCO a \href{https://microsoft.github.io/msmarco/}{\textit{Non-commercial research purposes only}}, NaturalQuestions a \href{https://ai.google.com/research/NaturalQuestions/download}{\textit{Creative Commons Share-Alike 3.0}} and the license for Quora follows their \href{https://www.quora.com/about/tos}{terms of service}. All the above allow the re-publishing of data for non-commercial purposes, hence we release the full dataset, complemented by our annotations. 

\paragraph{Risk of Misuse}
We do not see direct potential for misuse or harm due to \ourdata. The intended use is to spur research about human curiosity and information-seeking behavior. All the data sources were already available before this work. The insights about human behavior should be taken into consideration with care, considering their possible limitations. 

Also, since some of the sources used to build \ourdata may contain toxic and/or NSFW samples, we advise users that \ourdata might necessitate filtering and pre-processing to ensure the safety and appropriateness of the dataset for downstream applications.

\ifarxiv
\section*{Author Contributions}\label{sec:contributions}
The \textit{idea and design} of the project originated in discussions among Zhijing, Dmitrii, Roberto, and refined with Mrinmaya. In the initial exploration stage, Dmitrii started collecting the data and studying the questions in detail to find meaningful categorizations. The \textit{design of the taxonomies} was developed in discussions among Roberto and Zhijing. The \textit{data collection} (including prompt engineering, manual annotation, experimentation with GPT and labelling of \ourdata) was led by Roberto, with the important contribution of Ahmad and Amélie. The \textit{fine-tuning} of the language models was led by Dmitrii. The \textit{cleaning and compilation} of the code and data was done by Roberto.  The \textit{final-stage revision and formatting} was done by Punya. All co-authors contributed to writing the paper, especially Roberto, Zhijing and Ahmad, with significant feedback from Mrinmaya, Rada and Bernhard.

\section*{Acknowledgment}
This material is based in part upon work supported by the German Federal Ministry of Education and Research (BMBF): Tübingen AI Center, FKZ: 01IS18039B; by the Machine Learning Cluster of Excellence, EXC number 2064/1 – Project number 390727645; by a National Science Foundation award (\#2306372); by a Swiss National Science Foundation award (\#201009) and a Responsible AI grant by the Haslerstiftung.
The usage of OpenAI credits are largely supported by the Tübingen AI Center.
Zhijing Jin is supported by PhD fellowships from the Future of Life Institute and Open Philanthropy, as well as travel support from ELISE (GA \#951847) for the ELLIS program. 
\fi

\bibliography{sec/refs_zhijing,sec/refs_causality,sec/refs_cogsci,sec/refs_nlp4sg,sec/refs_semantic_scholar,refs}
\bibliographystyle{acl_natbib}

\clearpage
\appendix

\section{Additional Analyses}
\label{app:analysis}

\begin{table}[ht]
\centering \small
\begin{tabular}{lc}
\toprule
{Aspect} & {Cohen's Kappa} \\
\midrule
Cognitive complexity & 0.6926 (weighted) \\
Users' needs & 0.58 \\ 
Open-endedness & 0.63 \\
Domain & 0.73 \\
Subjectivity & 0.68 \\
\bottomrule
\end{tabular}
\label{tab:reliability_measures}
\caption{Inter-rater reliability measures for different aspects of question evaluation.}
\end{table}

\subsection{Cluster Details}
\label{app:cluster}
In \cref{tab:cluster-examples} we show some question examples from each cluster in \ourdata.

\begin{table*}[h!]
\centering \small
\label{tab:cluster-examples}
\setlength\tabcolsep{3pt}
\begin{tabular}{p{2cm}p{13cm}}
\toprule
{Cluster} & {Example Questions} \\
\midrule
Daily life & 
\begin{itemize}[nosep]
     \item How can I marry a millionaire? 
     \item What should I say when someone is expressing concern for my health?
     \item How can I cope with feeling constantly numb and unmotivated?
     \item Can you suggest anchor thoughts for optimizing my mindset, especially in the morning?
     \item Should I follow up with him to ask about his feelings, even if it may result in rejection?
\end{itemize} \\ \hline
Computer-Related & 
\begin{itemize}[nosep]
    \item Why are the signals generated by my code incorrect?
    \item Can you create a 3D wireframe grid in JavaScript on a canvas without frameworks or libraries, allowing for snapping lines and adding points?
    \item How can I explain to the brand/marketing team that iOS app deployment may take up to 24 hours, causing updates to not be instantly available to all customers?
    \item What is the use of static keyword in Java?
    \item Can I use the rules and the design (not logo) of monopoly in my mobile game?
\end{itemize} \\ \hline
Sports, medicine and science & 
\begin{itemize}[nosep]
    \item Where does the fertilization of the egg occur?
    \item What type of image is formed by a 4 cm object placed 40 cm away from a convex mirror with a radius of curvature of 20 cm?
    \item A decrease in the normal amount of urine is called oliguria.
    \item List of high school players drafted in NBA.
    \item Which hormone is most responsible for signaling satiety as well as reducing food intake during a meal?
\end{itemize} \\ \hline
Prompt Questions & 
\begin{itemize}[nosep]
    \item Generate detailed image prompts for the AI ``Midjourney'' based on given concepts, varying in description, environment, composition, atmosphere, and style.
    \item Generate detailed image prompts for the AI ``Midjourney'' in Chinese ink style, depicting a poet alone in the snow, smelling plum blossoms from a wine glass.
    \item Can you generate imaginative prompts for Midjourney's AI program to inspire unique and interesting images? Start with something beyond human imagination.
    \item Generate detailed image prompts for the AI ``Midjourney'' featuring a concept, environment, composition, mood, and style. Follow the specified structure and guidelines.
\end{itemize} \\ \hline
Stories and Fictional characters & 
\begin{itemize}[nosep]
    \item Where does the new Beauty and the Beast take place?
    \item How will Game of Thrones end?
    \item The Legend of Zelda: Why does Link have pointy ears?
    \item Short summary of Miss Peregrine's Home for Peculiar Children book.
    \item Creatively name and describe fruits and vegetables from an alien world
\end{itemize} \\
\bottomrule
\end{tabular}
\caption{Example questions for each topic cluster.}
\end{table*}

\begin{table}[h]
\centering \small
\label{tab:clusters}
\setlength\tabcolsep{3pt}
\begin{tabular}{p{2cm}ccccc}
\toprule
{Cluster} & {\# Questions} & {Causal (\%)} & {Non-Causal (\%)} \\
\midrule
Daily life & 3,609 & 56.08 & 43.92 \\ \hline
Computer-Related & 3,806 & 59.88 & 40.12 \\ \hline
Sports, medicine and science & 3,875 & 22.99 & 77.01 \\ \hline
Prompt Questions & 216 & 32.41 & 67.59 \\ \hline
Stories and Fictional characters & 1,994 & 21.92 & 78.08 \\ 
\bottomrule
\end{tabular}
\caption{Causal and non-causal questions in each topic cluster.}
\end{table}
\vspace{0.5em}

\subsection{Data Statistics for Causal and Non-Causal Questions}
\label{app:a2}

Here we show the overall statistics for causal and non-causal questions in \ourdata.

\begin{table}[H]
\centering \small
\begin{tabular}{lccc}
\toprule
& {Overall} & {Causal} & {Non-Causal} \\
\midrule
\# Samples & 13,500 & 5,701 & 7,799 \\
{\# Words/Sample} & 11.34 & 12.56 & 10.45 \\ 
Vocab Size & 25,709 & 14,526 & 17,367 \\
{Type-Token Ratio} & 0.168 & 0.202 & 0.213 \\
\bottomrule
\end{tabular}
\label{tab:stats_all}
\caption{Overall statistics of our \ourdata dataset.}
\end{table}

\subsection{Subjectivity}
\label{app:subjectivity}
Questions are also classified based on whether they are subjective (i.e., involving personalized, culture-specific opinions, etc.) or objective. We find that 
SE receives mostly objective questions, and users typically don’t raise too many subjective, opinion-seeking questions to them. H-to-H online forums receive 53\% subjective questions. LLMs receive 29\% subjective questions, although the majority are still objective, factual ones. There is still a way to go for LLMs to gain more trust and to let people ask more personally related questions.

\subsection{Politeness}
We look at politeness using the library ConvoKit \citep{chang2020convokit} and follow the rules to compute politeness introduced by \citet{danescu2013computational}. We build a composite \textit{politeness score} by adding $1$ for the presence of each positive linguistic indicator of politeness, and subtracting $1$ for negative ones. We refer to \citet{danescu2013computational} for more details about positive and negative markers. In line with intuition, we find that internet users are mostly polite with other humans (Quora). Interestingly, interactions with LLMs follow, and finally SEs. 

\subsection{Evaluation of Different Models on Causal Question Classification}
\label{app:causal-classifiers}
Here we train or fine-tune various efficient models for causal question classification.
\paragraph{A Selection of Efficient Models}
We first train a baseline model that does not involve deep learning, namely XGBoost \citep{chen2016xgboost} using TF-IDF for vectorization \citep{sparckjones1972statistical}, and we fine-tune 5 language models, with a number of parameters ranging from 80M to 2.85B \citep{chung2022scaling, li2023textbooks}.
We use full-weight supervised fine-tuning for smaller models and LoRA \citep{hu2022lora} for the largest ones to save on computing power.  For models running on 3090s RTX, the hardware has 24GB of memory, and the A100 has 40 GB. 

\cref{tab:training_results} highlights the tradeoff between the accuracy and size of the model. Future practitioners who may want to use \textit{causal classification} have several valid options, depending on the balance they need between resource usage and accuracy. For example, FLAN-T5-XL (LoRA) performed best but required significantly more resources than all other models. Smaller FLAN-T5 models and Phi-1.5 represent a compromise, giving up 2-3\% but saving on computations. Finally, using our baseline can also be an option for use cases with very limited computing power, being extremely fast and lightweight and still achieving a 72\% F1 score on the task. We also check the performance of a linguistics-based classifier, which integrates both lexical rules based on \citet{Bondarenko2022CausalQA} and lexico-syntactic patterns along with sets of causal and morphological connectives in \citet{girju2002mining}. This classifier achieves an accuracy of 23\% and a precision of 63\% in identifying causal questions. The main reason for its limited performance is that our curiosity-driven questions are often more formulated in an informal manner, contrasting with more standard grammar used in the other datasets. For example, the question word is implicit in the sample ``\textit{win a grand slam without losing a set?}''. These results suggest that effective identification of causal questions requires a deeper, semantic understanding of the questions as well as background knowledge, in particular, to be able to capture the implicit and handle the ambiguity intrinsic to many causative constructions \citep{girju2002mining}.

\subsection{GPT-4o evaluation on \ourdata}
\label{app:answer-eval-gpt4o}
In this section, we evaluate the performance of GPT-4o on \ourdata. 

\ourdata does not contain ground truth answers due to the open-ended nature of many of the questions (see \cref{app:subjectivity}). This is also the case for several of our data channels, in particular, both H-to-LLMs and H-to-H datasets don’t have answers. Among H-to-SEs datasets, NaturalQuestions have a Wikipedia passage containing the answer, and MSMarco has human-written answers.

Nevertheless, we can evaluate answers based on a metric of 3 usability criteria used in user satisfaction surveys \citep{international2018ergonomics}: \textit{effectiveness} (does the answer complete the goal specific to this user?), \textit{efficiency} (does the answer provide just the right amount of information or is it too vague / overly detailed?) and \textit{satisfaction} (is the answer friendly and pleasant to read, leaving the user satisfied?). 

We assess the answer quality on $50$ randomly sampled causal questions from \ourdata for GPT-4o \citep{gpt4o}. Two human annotators score the answers on a scale from 0 to 5 (similar to a Likert scale \citep{joshi2015likert}) on our metric defined above. We obtain a mean effectiveness of $3.83$ (std $0.97$), efficiency $2.88$ (std $0.78$) and satisfaction $3.85$ (std $0.63$). We find that GPT-4o struggles when asked to foresee the future or questions related to personal decisions, but performs well on causal questions requiring knowledge lookup.
The answers are also overly verbose (reflected in the lower efficiency score) and struggle with empathy and friendliness as the answers almost always just provide a list of factors to consider without engaging with the user. Tackling these weaknesses can be a large avenue for future work to achieve AI agents better equipped to get closer to the types of answers that humans seek.

\section{Prompts}
\label{app:prompts}
All prompts have a similar structure: each possible category is defined in detail, then some examples are provided. 

\subsection{Summarisation}
\begin{tcolorbox}[breakable, colback=customlightblue!100,colframe=customblue2!100, title=Needs, colbacktitle=customblue2!100]
    Below you’ll find a question that a human asked on \{source\}. Reformulate the question more concisely, while retaining the original key idea. You can skip the details. Maximum length: 30 words.\\
    Question: \{question\} \\
    Shorter question:
\end{tcolorbox}
\subsection{Causality}
\begin{tcolorbox}[breakable, colback=customlightblue!100,colframe=customblue2!100, title=Causality, colbacktitle=customblue2!100]

        The following is a question that a human asked online. Classify the question in one of the following two categories: 

        Category 1: Causal. This category includes questions that suggest a cause-and-effect relationship broadly speaking, requiring the use of cause-and-effect knowledge or reasoning to provide an answer.
        A cause is a preceding thing, event, or person that contributes to the occurrence of a later thing or event, to the extent that without the preceding one, the later one would not have occurred.
        A causal question can have different mechanistic natures. It can be:
        1. Given the cause, predict the effect: seeking to understand the impact or outcome of a specific cause, which might involve predictions about the future or hypothetical scenarios.\\
        2. Given the effect, predict the cause: asking ``Why'' something occurs (e.g. Why do apples fall?), probing the cause of a certain effect, asking about the reasons behind something, or the actions needed to achieve a specific result or goal, ``How to'' do something, explicitly or implicitly (e.g. Why does far right politics rise nowadays? How to earn a million dollars? How to learn a language in 30 days?). 
        This also includes the cases in which the effect is not explicit: any request with a purpose, looking for ways to fulfill it. It means finding the action (cause) to best accomplish a certain goal (effect), and the latter can also be implicit. If someone asks for a restaurant recommendation, what she’s looking for is the best cause for a certain effect which can be, e.g., eating healthy. If asking for a vegan recipe, she’s looking for the recipe that causes the best possible meal. Questions asking for ``the best'' way to do something, fall into this category. 
        Asking for the meaning of something that has a cause, like a song, a book, a movie, is also causal, because the meaning is part of the causes that led to the creation of the work. A coding task which asks for a very specific effect to be reached, is probing for the cause (code) to obtain that objective. \\
        3. Given variables, judge their causal relation: questioning the causal link among the given entities (e.g. Does smoking cause cancer? Did my job application get rejected because I lack experience?)\\
        Be careful: causality might also be implicit! Some examples of implicit causal questions: 
        - the best way to do something\\
        - how to achieve an effect\\
        - what’s the effect of an action (which can be in the present, future or past)\\
        - something that comes as a consequence of a condition (e.g. how much does an engineer earn, what is it like to be a flight attendant)\\
        - when a certain condition is true, does something happen?\\
        - where can I go to obtain a certain effect?
        - who was the main cause of a certain event, author, inventor, founder?\\
        - given an hypothetical imaginary condition, what would be the effect?\\
        - what’s the feeling of someone after a certain action?\\
        - what’s the code to obtain a certain result?\\
        - when a meaning is asked, is it because an effect was caused by a condition (what’s the meaning of <effect>)?\\
        - the role, the use, the goal of an entity, an object, is its effect\\

        Category 2: Non-causal. This category encompasses questions that do not imply in any way a cause-effect relationship.

        Let's think step by step.
        Question: \{question\}
\end{tcolorbox}

\subsection{Cognitive Complexity}
We follow a commonly used taxonomy for evaluating the kinds of intellectual skills needed to answer a question by \citet{anderson2001taxonomy}, using GPT4o-mini to classify questions into the different required skills. We follow a procedure similar to \cref{sec:prompt_improvement} to validate the efficacy of our prompt. We further evaluated our results against the group truth, achieving a $0.64$ Cohen's kappa score \citep{banerjee1999beyond, ullrich2009pedagogically}, which is interpreted as ``moderate agreement.''

\begin{tcolorbox}[breakable, colback=customlightblue!100,colframe=customblue2!100, title=Cognitive Complexity, colbacktitle=customblue2!100]
Your task is to classify given statements or questions according to Anderson and Krathwohl’s Taxonomy of the Cognitive Domain. Use the following six categories and their descriptions:

Remembering: Recognizing or recalling knowledge from memory. Remembering is when memory is used to produce or retrieve definitions, facts, or lists, or to recite previously learned information. Factual questions, that do not require reasoning fall into this category. \\
Understanding: Constructing meaning from different types of functions be they written or graphic messages or activities like interpreting, exemplifying, classifying, summarizing, inferring, comparing, or explaining. Questions asking for the meaning or explanation of a concept fall into this category.\\
Applying: Carrying out or using a procedure through executing, or implementing. Applying relates to or refers to situations where learned material is used, applied in a concrete situation, is used to present or show something. Questions that require the application of some theory or rule. For example, requiring some calculation, formula, light reasoning, applied to something in the real world. They do not entail a creative effort, but instead applying some rule or principle. Asking to generate a code with a specific goal (e.g. a cmd code that does yyy) is ``apply'' whereas asking to build a website requires a creative effort. How to do something, how to make something, how to solve something, how to apply some principle etc.\\
Analyzing: Breaking materials or concepts into parts, determining how the parts relate to one another, or how the parts relate to an overall structure or purpose. Mental actions included in this function are differentiating, organizing, and attributing, as well as being able to distinguish between the components or parts. When one is analyzing, he/she can illustrate this mental function by creating spreadsheets, surveys, charts, or diagrams, or graphic representations. Questions requiring deeper, more complex considerations on a certain thing. e.g. considering several aspects of something, considering pros and cons etc. Explaining why something is the way it is, by providing evidence or logical reasoning.\\
Evaluating: Making judgments based on criteria and standards through checking and critiquing. Critiques, recommendations, and reports are some of the products that can be created to demonstrate the processes of evaluation. Evaluating comes before creating as it is often a necessary part of the precursory behavior before one creates something. Questions asking to make judgements, suggestions, recommendations. Also making an hypothesis about something uncertain. Judging whether something is better than something else, or the best. \\
Creating: Putting elements together to form a coherent or functional whole; reorganizing elements into a new pattern or structure through generating, planning, or producing. Creating requires users to put parts together in a new way, or synthesize parts into something new and different creating a new form or product. This process is the most difficult mental function in the taxonomy. Questions asking for generation tasks that require a creative effort fall into this category. \\

Examples

1. Q: What does the term 'photosynthesis' mean?\\
    Classification: Understanding\\
    Explanation: The question asks for the meaning of a term, which falls under the 'Understanding' category.\\
2. Q: Calculate the area of a circle with a radius of 5 meters.\\
    Classification: Applying\\
    Explanation: The question requires applying a formula to calculate the area of a circle, which falls under the 'Applying' category.\\
3. Q: Compare and contrast the advantages and disadvantages of renewable energy sources.\\
    Classification: Evaluating\\
4. Q: Design a new logo for a tech startup company.\\
    Classification: Creating\\
5. Q: Explain the causes of World War II.\\
    Classification: Analyzing\\
6. Q: What category does the word 'dog' belong to?\\
    Classification: Remembering\\
7. Q: Is surfing easier to learn than snowboarding?\\
    Classification: Evaluating\\

Please classify the following question:\\

\{question\}
\end{tcolorbox}

\subsection{Knowledge Domains}
To classify \ourdata into Knowledge Domains, we use an iterative category generation procedure. We begin with an initial categorization, classify 1000 points using GPT-3.5-turbo-0125 including an ``Other'' category, and finally manually inspect the ``Other'' category to refine the categories. This procedure is repeated until a satisfactory categorization is achieved.
Based on the categories, we adopt the iterative   prompting improvement to optimize the prompt according to the human annotated set.
\begin{tcolorbox}[breakable, colback=customlightblue!100,colframe=customblue2!100, title=Domain, colbacktitle=customblue2!100]
Below you’ll find a question. Classify it in one of the following categories:

1. Natural and Formal Sciences: This category encompasses questions related to the physical world and its phenomena, including, but not limited to, the study of life and organisms (Biology), the properties and behavior of matter and energy (Physics), and the composition, structure, properties, and reactions of substances (Chemistry); also formal sciences belong to this category, such as Mathematics and Logic. Questions in this category seek to understand natural laws, the environment, and the universe at large.\\
2. Society, Economy, Business: Questions in this category explore the organization and functioning of human societies, including their economic and financial systems. Topics may cover Economics, Social Sciences, Cultures and their evolution, Political Science and Law. Questions regarding business, sales, companies’ choices and governance fall into this category. \\
3. Health and Medicine: This category focuses on questions related to human health, diseases, and the medical treatments used to prevent or cure them. It covers a wide range of topics from the biological mechanisms behind diseases, the effectiveness of different treatments and medications, to strategies for disease prevention and health promotion. It comprises anything related or connected to human health.\\
4. Computer Science and Technology: Questions in this category deal with the theoretical foundations of information and computation, along with practical techniques for the implementation and application of these foundations. Topics include, but are not limited to, theoretical computer science, coding and optimization, hardware and software technology and innovation in a broad sense. This category includes the development, capabilities, and implications of computing technologies.\\
6. Psychology and Behavior: This category includes questions about the mental processes and behaviors of humans. Topics range from understanding why people engage in certain behaviors, like procrastination, to the effects of social factors, and the developmental aspects of human psychology, such as language acquisition in children. The focus is on understanding the workings of the human mind and behavior in various contexts, also in personal lives.\\
7. Historical Events and Hypothetical Scenarios: This category covers questions about significant past events and their impact on the world, as well as hypothetical questions that explore alternative historical outcomes or future possibilities. Topics might include the effects of major wars on global politics, the potential consequences of significant historical events occurring differently, and projections about future human endeavors, such as space colonization. This category seeks to understand the past and speculate on possible futures or alternative historical happenings.\\
8. Everyday Life and Personal Choices: Questions in this category pertain to practical aspects of daily living and personal decision-making. Topics can range from career advice, cooking tips, and financial management strategies to advice on maintaining relationships and organizing daily activities. This category aims to provide insights and guidance on making informed choices in various aspects of personal and everyday life. Actionable tips fall into this category. \\
9. Arts and Culture: This category includes topics in culture across various mediums such as music, television, film, art, games, and social media, sports, celebrities. \\

Assign one of the above categories to the given question. 

Question: \{question\}

\end{tcolorbox}

\subsection{Users' needs}
\label{app:needs}

For user needs, LLMs reach an F1 score of $0.80$ with corresponding human annotations.

\begin{tcolorbox}[breakable, colback=customlightblue!100,colframe=customblue2!100, title=Needs, colbacktitle=customblue2!100]
Analyze the following question and identify the primary user need category it falls into. Consider the broad categories of user needs as defined below:

Knowledge and Information: Seeking factual information, understanding concepts, or exploring ideas. Questions falling in this category are looking for knowledge for its own sake, as far as we can infer from the question. We do not see an underlying need of the user except for curiosity.\\
Problem-Solving and Practical Skills: Troubleshooting issues, learning new skills, managing daily life, and handling technology. Anything that is actionable, how to do something or solve a problem. \\
Personal Well-being: Improving mental and physical health, managing finances, and seeking support.\\
Professional and Social Development: Advancing career, job searching, and improving social interactions. Any information request about work, school, academia (but that is not actionable, in that case it’s 2. Problem-Solving and Practical Skills). \\
Leisure and Creativity: Finding recreational activities, pursuing hobbies, and seeking creative inspiration.\\
Categorize this question based on the user's primary need asked in the question, choosing among the categories above. 

Examples

1. Q: ``What is the chemical symbol for gold?''\\
    A: Knowledge and Information\\
2. Q: ``How do I fix a leaky faucet?''\\
    A: Problem-Solving and Practical Skills\\
3. Q: ``What does the paracetamol do to the body?''\\
    A: Personal Well-being\\
4. Q: ``Create a story about a magical kingdom.'' \\
    A: Leisure and Creativity\\
5. Q: ``How do I improve my resume?''\\
    A: Professional and Social Development\\

Question: ``{question}''

\end{tcolorbox}

\subsection{Open-Endedness}
\label{app:b-open-endedness}
\begin{tcolorbox}[breakable, colback=customlightblue!100,colframe=customblue2!100, title=Open-Endedness, colbacktitle=customblue2!100]
You are tasked with classifying answerable questions based on the uniqueness of their answers. This classification helps understand the nature of the question and the potential diversity of valid responses.

For answerable questions:
1. Unique Answer: There is a single, specific correct answer that is widely accepted based on current human knowledge. Questions about fictional characters are most likely ``Unique answer'' because most of the times we can assume the answer is in the book / movie \\
2. Multiple Valid Answers: There are several plausible, valid answers that could be considered correct depending on perspective, context, or interpretation. Multiple valid answers means either (1) there is a subjective judgment involved (the answer varies depending on who answers), or (2) it is a creative task that can be solved in several different ways, or (3) we as humans do not have access to a unique correct answer. Trivial different wordings of the same concept do not qualify as ``Multiple answers''. e.g. ``What's the meaning of bucolic?'' \\

Ambiguous questions are not necessarily ``multiple answers'' just because they are ambiguous (i.e. since different people could understand the query differently, they might give different answers). In such cases we should assume a meaning for the question, and reason about the possible ways to answer it.

For each given, classify it as either:
1. Unique Answer\\
2. Multiple Valid Answers\\

If the question is looking for a list of items, but the list is unique and well defined, it should be classified as ``Unique Answer''. \\
If the question is looking for a number which can be found or computed, it should be classified as ``Unique Answer''.\\
If the question is asking how something can be achieved, and there is only one way to achieve it, it should be classified as ``Unique Answer'', and ``Multiple Valid Answers'' otherwise.

Examples

1. Q: ``What is the chemical symbol for gold?''\\
  Classification: Unique Answer\\
  Explanation: There is a single, universally accepted answer in chemistry: Au.\\

2. Q: ``What is the best programming language for web development?''\\
  Classification: Multiple Valid Answers\\
  Explanation: There are several programming languages suitable for web development, and the ``best'' can depend on project requirements, developer preference, and other factors.\\

3. Q: ``Who was the first president of the United States?''\\
  Classification: Unique Answer\\
  Explanation: There is a single, historically accepted answer: George Washington.\\

4. Q: ``Tell me some short bedtime stories''\\
  Classification: Multiple Valid Answers\\
  Explanation: There are various short stories that can be told at bedtime, and the choice can vary based on cultural background, personal preference, and other factors.\\

5. Q: ``What is the meaning of life?''\\
    Classification: Multiple Valid Answers\\
    Explanation: This question has multiple valid answers based on different philosophical, religious, and personal perspectives.\\

6 Q: ``What does it mean when an economy is in a recession?''\\
    Classification: Unique Answer\\
    Explanation: There is a specific definition of a recession in economics, making this a question with a unique answer.\\

7 Q: ``Name the three primary colors''\\
    Classification: Unique Answer\\
    Explanation: There are three primary colors in the RGB color model: red, green, and blue.\\

8 Q: ``What criteria should I consider when buying a new laptop?''\\
    Classification: Multiple Valid Answers\\
    Explanation: The criteria for buying a laptop can vary based on individual needs, preferences, and budget constraints.\\

9. Q: ``Who is the best neurosurgeon in New York?''\\
    Classification: Multiple Valid Answers\\
    Explanation: The best neurosurgeon can vary based on specialization, patient reviews, and other factors.

Please classify the following question:\\

\{question\}

\end{tcolorbox}

\subsection{Prompt Iteration}
\label{sec:prompt_iteration_examples}
Here we show the first and the last prompts of the iteration procedure for Causality and Open-Endedness. 

\subsubsection{Causality}
With first prompt we obtained a weighted F1 of 0.798, with the sixth iteration we obtained 0.894. 

\begin{tcolorbox}[breakable, colback=customlightblue!100,colframe=customblue2!100, title=First Prompt Causality, colbacktitle=customblue2!100]
The following is a question that a human asked on {website}. Classify the question in one of the following two categories:

Category 1: Causal. This category includes questions that suggest a cause-and-effect relationship broadly speaking, requiring the use of cause-and-effect knowledge or reasoning to provide an answer.

A causal question can have different mechanistic natures. It can be:
1. Given the cause, predict the effect: seeking to understand the impact or outcome of a specific cause, which might involve predictions about the future or hypothetical scenarios (e.g. What if I do a PhD? Should I learn how to swim? Will renewable energy sources become the primary means of power? What would the world look like if the Internet had never been invented?);

2. Given the effect, predict the cause: asking ``Why'' something occurs (e.g. Why do apples fall?), probing the cause of a certain effect, asking about the reasons behind something, or the actions needed to achieve a specific result or goal, ``How to'' do something, explicitly or implicitly (e.g. Why does far right politics rise nowadays? How to earn a million dollars? How to learn a language in 30 days?).
This also includes the cases in which the effect is not explicit: any request with a purpose, looking for ways to fulfill it. It means finding the action (cause) to best accomplish a certain goal (effect), and the latter can also be implicit. If someone asks for a restaurant recommendation, what she’s looking for is the best cause for a certain effect which can be, e.g., eating healthy. If asking for a vegan recipe, she’s looking for the recipe that causes the best possible meal. Questions asking for “the best” way to do something, fall into this category;

3. Given variables, judge their causal relation: questioning the causal link among the given entities (e.g. Does smoking cause cancer? Did my job application get rejected because I lack experience?)

Categorical 2: Non-causal. This category encompasses questions that do not imply in any way a cause-effect relationship. For example a non-causal question can be asking: \\
To translate, rewrite, paraphrase a text \\
To generate a story\\
To play a game\\
To provide the solution for a mathematical expression, or a riddle requiring mathematical reasoning\\
To provide information about something (softwares, websites, materials, events, restaurants) or use such information to make a comparison, without much reasoning. This is non-causal because there is not a specific purpose of the user, but they are only looking for information.

Examples:
Question: What would the world look like if the Internet had never been invented?
Category: <Causal>

Question: I'd like to play a game of Go with you through text. Let's start with a standard 19x19 board. I'll take black. Place my first stone at D4.
Category: <Non-causal>

Question: How can I earn a million dollars fast?
Category: <Causal>

Question: How should I spend my last month in Argentina before leaving the country for a long time?
Category: <Causal>

Question: Translate “hiking” in Italian
Category: <Non-causal>

Question: Will renewable energy sources become the primary means of power?
Category: <Causal>

Question: What's the derivative of the logarithm?
Category: <Non-causal>

Question: What are some high-protein food options for snacks?
Category: <Non-causal>

Question: Does smoking cause cancer?
Category: <Causal>

Question: What's the best vegan recipe with broccoli?
Category: <Causal>

Question: Write a python script to efficiently sort an array.
Category: <Causal>

Question: What's more efficient, Python or C++?
Category: <Non-causal>

Question: Tell me the names of all bookshops in Zurich
Category: <Non-causal>

Question: Best chair for a home office
Category: <Causal>

Answer ONLY with the category in the following format: <Category>, e.g. <Causal>, <Non-causal>.

Question: {question}
Category:

\end{tcolorbox}

\begin{tcolorbox}[breakable, colback=customlightblue!100,colframe=customblue2!100, title=Last Prompt Causality (6th iteration), colbacktitle=customblue2!100]
The following is a question that a human asked on \{website\}. Classify the question in one of the following two categories:

       Category 1: Causal. This category includes questions that suggest a cause-and-effect relationship broadly speaking, requiring the use of cause-and-effect knowledge or reasoning to provide an answer. \\
       A cause is a preceding thing, event, or person that contributes to the occurrence of a later thing or event, to the extent that without the preceding one, the later one would not have occurred.\\
       A causal question can have different mechanistic natures. It can be:
       1. Given the cause, predict the effect: seeking to understand the impact or outcome of a specific cause, which might involve predictions about the future or hypothetical scenarios.\\
       2. Given the effect, predict the cause: asking ``Why'' something occurs (e.g. Why do apples fall?), probing the cause of a certain effect, asking about the reasons behind something, or the actions needed to achieve a specific result or goal, ``How to'' do something, explicitly or implicitly (e.g. Why does far right politics rise nowadays? How to earn a million dollars? How to learn a language in 30 days?).\\
       This also includes the cases in which the effect is not explicit: any request with a purpose, looking for ways to fulfill it. It means finding the action (cause) to best accomplish a certain goal (effect), and the latter can also be implicit. If someone asks for a restaurant recommendation, what she’s looking for is the best cause for a certain effect which can be, e.g., eating healthy. If asking for a vegan recipe, she’s looking for the recipe that causes the best possible meal. Questions asking for ``the best'' way to do something, fall into this category.\\
       Asking for the meaning of something that has a cause, like a song, a book, a movie, is also causal, because the meaning is part of the causes that led to the creation of the work. A coding task which asks for a very specific effect to be reached, is probing for the cause (code) to obtain that objective.\\
       3. Given variables, judge their causal relation: questioning the causal link among the given entities (e.g. Does smoking cause cancer? Did my job application get rejected because I lack experience?)\\
       Be careful: causality might also be implicit! Some examples of implicit causal questions:
       - the best way to do something\\
       - how to achieve an effect\\
       - what’s the effect of an action (which can be in the present, future or past)\\
       - something that comes as a consequence of a condition (e.g. how much does an engineer earn, what is it like to be a flight attendant)\\
       - when a certain condition is true, does something happen?\\
       - where can I go to obtain a certain effect?\\
       - who was the main cause of a certain event, author, inventor, founder?\\
       - given an hypothetical imaginary condition, what would be the effect?\\
       - what’s the feeling of someone after a certain action?\\
       - what’s the code to obtain a certain result?\\
       - when a meaning is asked, is it because an effect was caused by a condition (what’s the meaning of <effect>)?\\
       - the role, the use, the goal of an entity, an object, is its effect\\

       Category 2: Non-causal. This category encompasses questions that do not imply in any way a cause-effect relationship.

       Let's think step by step. Answer in the following format:
       Reasoning: [Reasoning]
       Category: [causal / Non-causal]

       Always write ``Category'' before providing the final answer.

       Question: \{question\}

\end{tcolorbox}

\subsubsection{Open-Endedness}
With the first prompt we obtained a weighted F1 of 0.713, with the fourth iteration we obtained 0.795. 

\begin{tcolorbox}[breakable, colback=customlightblue!100,colframe=customblue2!100, title=First Prompt Open-Endedness, colbacktitle=customblue2!100]
You are tasked with classifying answerable questions based on the uniqueness of their answers. This classification helps understand the nature of the question and the potential diversity of valid responses.

For answerable questions:
1. Unique Answer: There is a single, specific correct answer that is widely accepted based on current human knowledge. Questions about fictional characters are most likely ``Unique answer'' because most of the time we can assume the answer is in the book / movie
2. Multiple Valid Answers: There are several plausible, valid answers that could be considered correct depending on perspective, context, or interpretation. Multiple valid answers means either (1) there is a subjective judgment involved (the answer varies depending on who answers), or (2) it is a creative task that can be solved in several different ways, or (3) we as humans do not have access to a unique correct answer. Trivial different wordings of the same concept do not qualify as ``Multiple answers''. e.g. ``What's the meaning of bucolic?''.

Ambiguous questions are not necessarily ``multiple answers'' just because they are ambiguous (i.e. since different people could understand the query differently, they might give different answers). In such cases we should assume a meaning for the question, and reason about the possible ways to answer it.

For each given hypothetically answerable question, classify it as either:
1. Unique Answer
2. Multiple Valid Answers

Examples

1. Q: ``What is the chemical symbol for gold?''
 Classification: Unique Answer
 Explanation: There is a single, universally accepted answer in chemistry: Au.

2. Q: ``What is the best programming language for web development?''
 Classification: Multiple Valid Answers
 Explanation: There are several programming languages suitable for web development, and the ``best'' can depend on project requirements, developer preference, and other factors.

3. Q: ``Who was the first president of the United States?''
 Classification: Unique Answer
 Explanation: There is a single, historically accepted answer: George Washington.

4. Q: ``What is the most effective way to reduce stress?''
 Classification: Multiple Valid Answers
 Explanation: There are various effective stress reduction techniques, and the most effective method can vary from person to person.

Please classify the following question:

{question}

\end{tcolorbox}

\begin{tcolorbox}[breakable, colback=customlightblue!100,colframe=customblue2!100, title=Last Prompt Open-Endedness (4th iteration), colbacktitle=customblue2!100]

You are tasked with classifying answerable questions based on the uniqueness of their answers. This classification helps understand the nature of the question and the potential diversity of valid responses.

For answerable questions:
1. Unique Answer: There is a single, specific correct answer that is widely accepted based on current human knowledge. Questions about fictional characters are most likely ``Unique answer'' because most of the times we can assume the answer is in the book / movie
2. Multiple Valid Answers: There are several plausible, valid answers that could be considered correct depending on perspective, context, or interpretation. Multiple valid answers means either (1) there is a subjective judgment involved (the answer varies depending on who answers), or (2) it is a creative task that can be solved in several different ways, or (3) we as humans do not have access to a unique correct answer. Trivial different wordings of the same concept do not qualify as ``Multiple answers''. e.g. ``What's the meaning of bucolic?''

Ambiguous questions are not necessarily ``multiple answers'' just because they are ambiguous (i.e. since different people could understand the query differently, they might give different answers). In such cases we should assume a meaning for the question, and reason about the possible ways to answer it.

For each given hypothetically answerable question, classify it as either:
1. Unique Answer
2. Multiple Valid Answers

If the question is looking for a list of items, but the list is unique and well-defined, it should be classified as ``Unique Answer''.
If the question is looking for a number which can be found or computed, it should be classified as ``Unique Answer''.
If the question is asking how something can be achieved, and there is only one way to achieve it, it should be classified as ``Unique Answer'', and ``Multiple Valid Answers'' otherwise.

Examples

1. Q: ``What is the chemical symbol for gold?''
 Classification: Unique Answer
 Explanation: There is a single, universally accepted answer in chemistry: Au.

2. Q: ``What is the best programming language for web development?''
 Classification: Multiple Valid Answers
 Explanation: There are several programming languages suitable for web development, and the ``best'' can depend on project requirements, developer preference, and other factors.

3. Q: ``Who was the first president of the United States?''
 Classification: Unique Answer
 Explanation: There is a single, historically accepted answer: George Washington.

4. Q: ``Tell me some short bedtime stories''
 Classification: Multiple Valid Answers
 Explanation: There are various short stories that can be told at bedtime, and the choice can vary based on cultural background, personal preference, and other factors.

5. Q: ``What is the meaning of life?''
   Classification: Multiple Valid Answers
   Explanation: This question has multiple valid answers based on different philosophical, religious, and personal perspectives.

6 Q: ``What does it mean when an economy is in a recession?''
   Classification: Unique Answer
   Explanation: There is a specific definition of a recession in economics, making this a question with a unique answer.

7 Q: ``Name the three primary colors''
   Classification: Unique Answer
   Explanation: There are three primary colors in the RGB color model: red, green, and blue.

8 Q: ``What criteria should I consider when buying a new laptop?''
   Classification: Multiple Valid Answers
   Explanation: The criteria for buying a laptop can vary based on individual needs, preferences, and budget constraints.

9. Q: ``Who is the best neurosurgeon in New York?''
   Classification: Multiple Valid Answers
   Explanation: The best neurosurgeon can vary based on specialization, patient reviews, and other factors.

Please classify the following question:

{question}

\end{tcolorbox}

\begin{table*}[h]
\centering \small
\begin{tabular}{lccc}
\toprule
\textbf{Cognitive Complexity} & \textbf{Overall} & \textbf{Non-Causal} & \textbf{Causal} \\
\midrule
Remembering    & 36.82\% & 56.75\% & 9.54\% \\
Understanding  & 13.47\% & 11.54\% & 16.11\% \\
Applying       & 13.54\% & 3.67\%  & \textbf{27.04\%} \\
Analyzing      & 8.82\%  & 5.04\%  & 13.98\% \\
Evaluating     & 13.74\% & 9.09\%  & \textbf{20.11\%} \\
Creating       & 13.62\% & 13.90\% & 13.23\% \\
\bottomrule
\end{tabular}
\caption{Distribution of cognitive complexity across overall dataset, non-causal questions, and causal questions. Significant increases in causal questions are highlighted in bold.}
\end{table*}

\begin{table*}[h]
\centering \small
\begin{tabular}{lcccc}
\toprule
\textbf{Cognitive Complexity} & \textbf{Overall} & \textbf{H-to-SEs} & \textbf{H-to-H} & \textbf{H-to-LLMs} \\
\midrule
Remembering    & 36.82\% & \textbf{76.49\%} & 19.42\% & 14.52\% \\
Understanding  & 13.47\% & 13.82\% & 15.27\% & 11.32\% \\
Applying       & 13.54\% & 4.07\%  & 18.73\% & 17.81\% \\
Analyzing      & 8.82\%  & 3.40\%  & 13.00\% & 10.05\% \\
Evaluating     & 13.74\% & 1.87\%  & \textbf{31.27\%} & 8.09\% \\
Creating       & 13.62\% & 0.36\%  & 2.31\%  & \textbf{38.20\%} \\
\bottomrule
\end{tabular}
\caption{Distribution of cognitive complexity across different data sources. Notable values are highlighted in bold.}
\end{table*}

\begin{table*}[h]
\centering \small
\begin{tabular}{lcc}
\toprule
\textbf{Category} & \textbf{Non-Natural Questions} & \textbf{\ourdata} \\
\midrule
\multicolumn{3}{l}{\textit{\textbf{Domain Class}}}\\
Natural and Formal Sciences                      & 40.68\% & 7.74\% \\
Society, Economy, Business                       & 28.64\% & 18.79\% \\
Everyday Life and Personal Choices               & 7.74\%  & \textbf{16.00\%} \\
Computer Science                                 & 6.64\%  & \textbf{25.83\%} \\
Historical Events and Hypothetical Scenarios     & 5.90\%  & 4.48\% \\
Arts and Culture                                 & 3.83\%  & \textbf{9.41\%} \\
Health and Medicine                              & 3.77\%  & \textbf{10.91\%} \\
Psychology and Behavior                          & 2.80\%  & \textbf{6.83\%} \\
\hline
\multicolumn{3}{l}{\textit{\textbf{Bloom Taxonomy}}}\\
Applying        & \textbf{36.41\%} & 13.54\% \\
Remembering     & 30.51\% & 36.82\% \\
Analyzing       & 13.90\% & 8.82\%  \\
Evaluating      & 11.54\% & 13.74\% \\
Understanding   & 7.47\%  & 13.47\% \\
Creating        & 0.17\%  & \textbf{13.62\%} \\
\hline
\multicolumn{3}{l}{\textit{\textbf{Open-Endedness}}}\\
Unique Answer             & \textbf{70.43\%} & 32.30\% \\
Multiple Valid Answers    & 29.57\% & \textbf{67.70\%} \\
\hline
\multicolumn{3}{l}{\textit{\textbf{User Needs}}}\\
Knowledge and Information               & \textbf{73.70\%} & 57.09\% \\
Problem-Solving and Practical Skills    & 24.63\% & 14.77\% \\
Professional and Social Development     & 0.70\%  & \textbf{8.42\%} \\
Personal Well-being                     & 0.60\%  & \textbf{5.88\%} \\
Leisure and Creativity                  & 0.37\%  & \textbf{13.84\%} \\
\bottomrule
\end{tabular}
\caption{Comparison of domain classes, cognitive complexity, open-endedness, and user needs between curated questions and \ourdata. Significant differences are highlighted in bold.}
\end{table*}

\begin{table}[h]
\centering \small
\begin{tabular}{lcc}
\toprule
\textbf{Source} & \textbf{Non-Causal} & \textbf{Causal} \\
\midrule
Overall         & 57.77\% & 42.23\% \\
H-to-SEs        & \textbf{78.07\%} & 21.93\% \\
H-to-H          & 41.29\% & \textbf{58.71\%} \\
H-to-LLMs       & 53.96\% & 46.04\% \\
\bottomrule
\end{tabular}
\caption{Distribution of causal and non-causal questions across different data sources. Notable values are highlighted in bold.}
\end{table}

\section{Experimental Details}
\subsection{OpenAI Experiments and Labeling}
The total amount spent on the OpenAI API was around 200 dollars. When possible, the BatchAPI was used, to save on expenses. 
Several models were used across this work, depending on the availability at the moment of each analysis, and 

\begin{table}[h]
\centering \small
\begin{tabular}{lc}
\toprule
{Task} & {GPT Model} \\
\midrule
Summarisation & gpt-3.5-turbo-0125 \\
Causality classification & gpt-4-turbo-2024-04-09 \\
All other classifications & gpt-4o-mini-2024-07-18 \\
Sect 5.1 & gpt-4o-2024-05-13 \\
\bottomrule
\end{tabular}
\label{tab:gpt_models}
\caption{GPT models used for various tasks.}
\end{table}

Hyperparameters: 
\begin{itemize}
    \item \textit{seed}: 42
    \item \textit{temperature}: 1
    \item \textit{max tokens}:  1000
\end{itemize}

We used the OpenAI API endpoint for prompt engineering, testing, and gathering answers, and the BatchAPI to label the full dataset. 

\subsection{Trainings}
\label{sec:experimental_details}
\subsection*{Learning Rate}
The learning rate was always set to $2 \times 10^{-4}$.

\subsection*{FLAN-T5 LoRA Configuration}
\begin{lstlisting}[caption=FLAN-T5 LoRA Configuration]
lora_config = LoraConfig(
    r=32,
    lora_alpha=32,
    target_modules=["q", "v"],
    lora_dropout=0.05,
    bias="none",
    task_type=TaskType.SEQ_2_SEQ_LM  
)
\end{lstlisting}

\subsection*{PHI LoRA Configuration}
\begin{lstlisting}[caption=PHI LoRA Configuration]
lora_config = LoraConfig(
    r=32,
    lora_alpha=16,
    target_modules="all-linear",
    lora_dropout=0.05,
    bias="none",
    task_type=TaskType.CAUSAL_LM
)
\end{lstlisting}

\subsection*{Batch Size and Accumulation Details}
\begin{itemize}
    \item \textbf{PHI:} Batch size = 4, Accumulation = 1
    \item \textbf{XL:} Batch size = 1, Accumulation = 4
    \item \textbf{Large:} Batch size = 1, Accumulation = 4
    \item \textbf{Base:} Batch size = 1, Accumulation = 4
    \item \textbf{Small:} Batch size = 4, Accumulation = 1
\end{itemize}

\subsection*{Additional plots}
We report here additional plots on the performance of the fine-tuned classifiers. 

\begin{figure}[h]
    \centering
    \includegraphics[width=\linewidth]{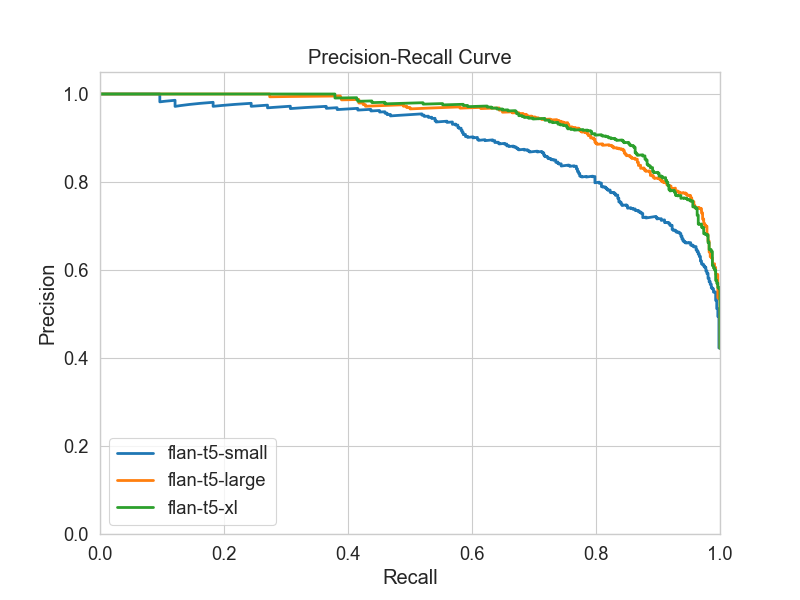}
    \caption{Combined Precision - Recall curve for the FLAN models.}
    
\end{figure}

\begin{figure}[h]
    \centering
    \includegraphics[width=\linewidth]{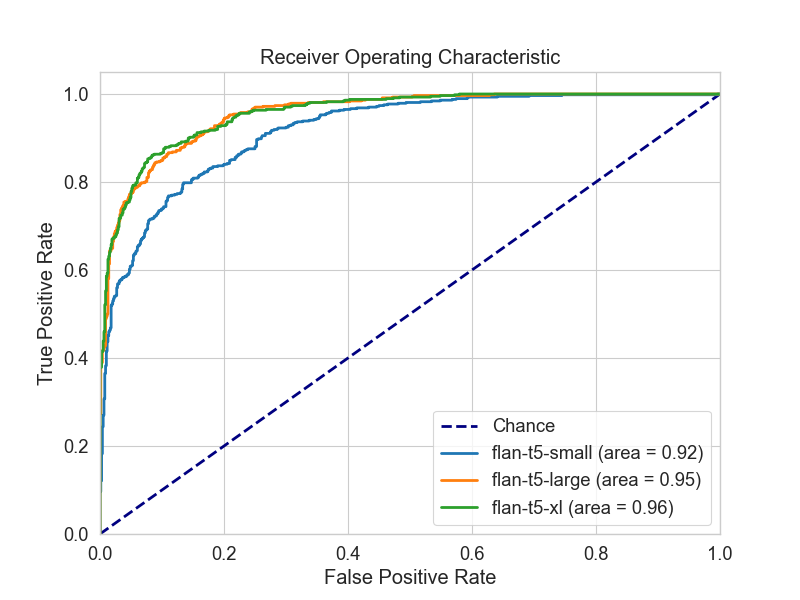}
    \caption{Combined ROC curve for the FLAN models.}
    
\end{figure}

\begin{figure}[h]
    \centering
    \includegraphics[width=\linewidth]{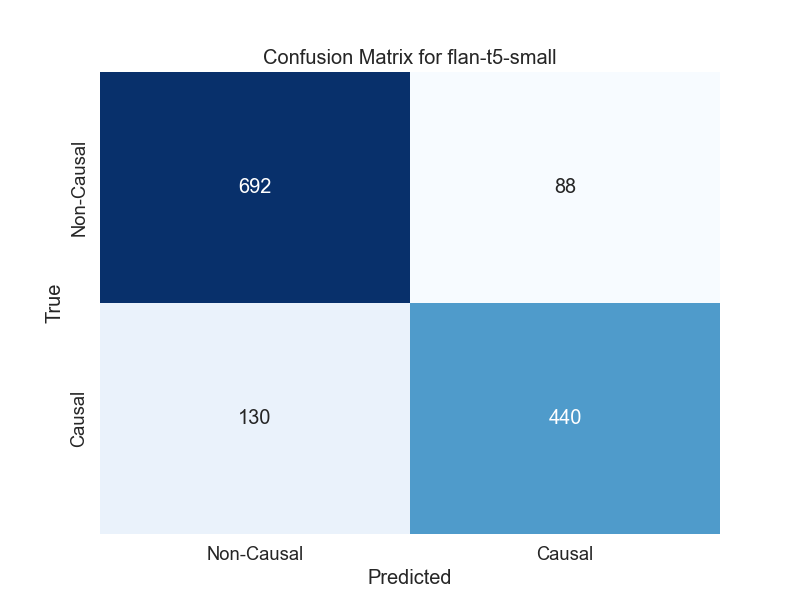}
    \caption{Confusion matrix for FLAN-T5-Small.}
    
\end{figure}

\begin{figure}[h]
    \centering
    \includegraphics[width=\linewidth]{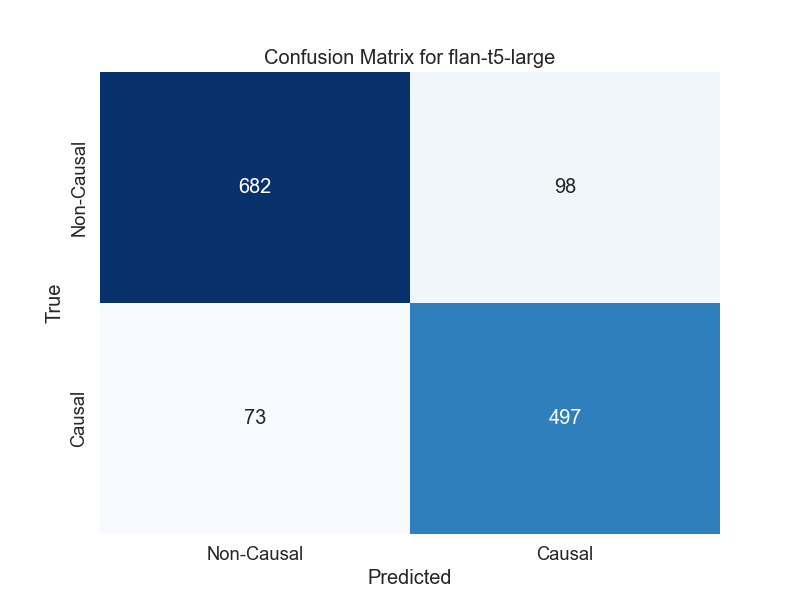}
    \caption{Confusion matrix for FLAN-T5-Large.}
    
\end{figure}

\begin{figure}[h]
    \centering
    \includegraphics[width=\linewidth]{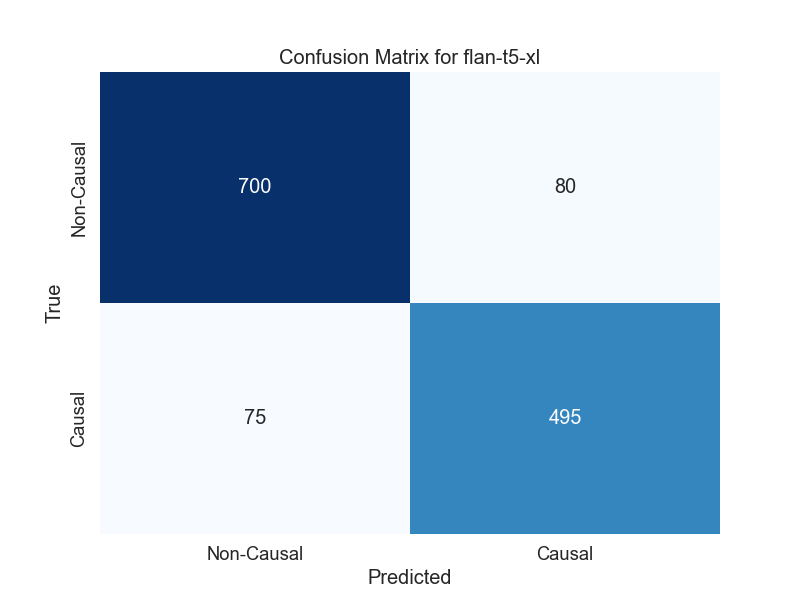}
    \caption{Confusion matrix for FLAN-T5-XL.}
    
\end{figure}

\begin{figure}[h]
    \centering
    \includegraphics[width=\linewidth]{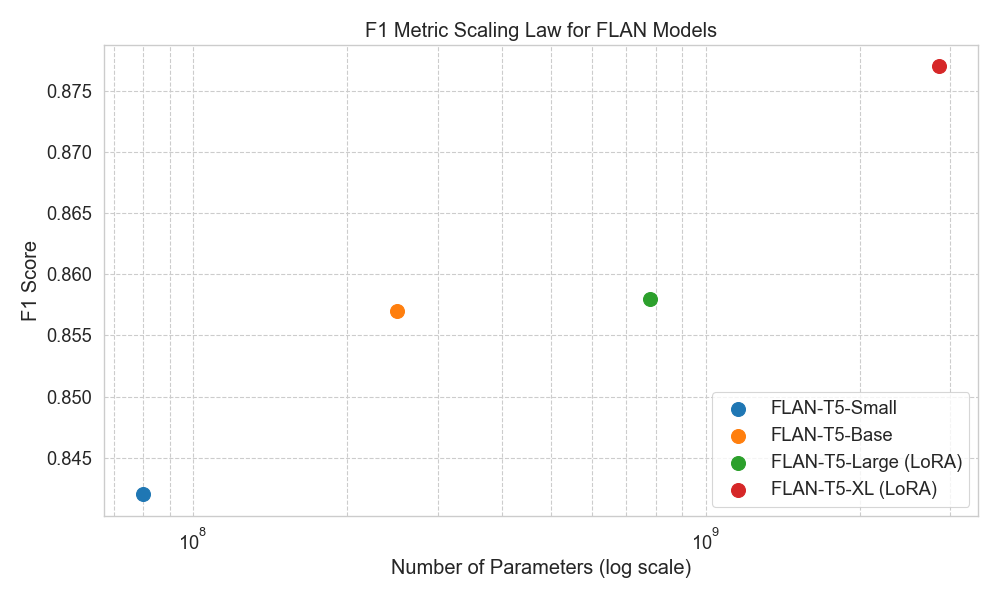}
    \caption{F1 score scaling law, FLAN models.}
    
\end{figure}

\section{More examples}

Within this section, we show some additional examples from \ourdata.
\subsection{Domain Classification}
\cref{tab:domain_examples} shows more examples classified by domain. 
\clearpage
\begin{table*}[htbp]
    \centering \small
    \begin{tabular}{p{0.25\linewidth} | p{0.75\linewidth}}
\toprule
        Domain Class &  Question \\ \midrule
        \multicolumn{2}{c}{Domain Class: Everyday Life and Personal Choices} \\ \midrule
        Everyday Life and Personal Choices & How can I go about finding an overseas T-shirt manufacturer for my clothing business?\\
        Everyday Life and Personal Choices & How do I care for raw denim?\\ \midrule
        \multicolumn{2}{c}{Domain Class: Computer Science} \\ \midrule
        Computer Science & Why was the question ``What does Jimmy Wales think of Wikipedia'' merged with ``What does Jimmy Wales think of Wikipedia Redefined?''\\
        Computer Science & Where can I get the design of CSR implementation on Block RAM?\\ \midrule
        \multicolumn{2}{c}{Domain Class: Psychology and Behavior} \\ \midrule
        Psychology and Behavior & What does it mean when a girl says 'You're so affectionate.'?\\
        Psychology and Behavior & What do dreams signify?\\ \midrule
        \multicolumn{2}{c}{Domain Class: Health and Medicine} \\ \midrule
        Health and Medicine & Is it healthy to eat seedless fruits?\\
        Health and Medicine & Why do you have swollen lymph nodes and what is the medicine for it?\\ \midrule
        \multicolumn{2}{c}{Domain Class: Natural and Formal Sciences} \\ \midrule
        Natural and Formal Sciences & How is AC converted into DC?\\
        Natural and Formal Sciences & What are the different parametric study done on the cable stayed bridge for dynamic loading?\\ \midrule
        \multicolumn{2}{c}{Domain Class: Society, Economy Business} \\ \midrule
        Society, Economy Business & Why are people now more interested in pop music rather that the good old classic rock?\\
        Society, Economy, Business & Is it true that when purchasing a vehicle, if the cost is under 7000 you do not need to have full coverage insurance?\\ \midrule
        \multicolumn{2}{c}{Domain Class: Historical Events and Hypothetical Scenarios} \\ \midrule
        Historical Events and Hypothetical Scenarios & What if Swami Vivekanand didn't get his guru?\\
        Historical Events and Hypothetical Scenarios & Will Facebook shut down by the end of 2016?\\ \midrule
        \multicolumn{2}{c}{Domain Class: Other} \\ \midrule
        Other & What are some of the worst examples of plagiarism you've witnessed?\\
        Other & Is Sanskrit considered a religious language?\\
\bottomrule
    \end{tabular}
    \label{tab:domain_examples}
        \caption{Examples of various classifications of questions based on their Domain class.}

\end{table*}

\begin{table*}[t]
    \centering \small
    \begin{tabular}{p{0.15\linewidth} | p{0.85\linewidth}}
\toprule
        Is Subjective & Question \\ \midrule
        \multicolumn{2}{c}{Subjectivity: False} \\ \midrule
        False & What is the origin of ES-IS and IS-IS? Is there any connection to the OSI model?\\
        False & Why does Hilary Clinton cough so much?\\ \midrule
        \multicolumn{2}{c}{Subjectivity: True} \\ \midrule
        True & I got a BigData internship and I want to get an internship in a bigger company next year. What skills should I spend my time on this summer?\\
        True & I am in college and it seems like every guy ignores me. Is there something wrong with me?\\
\bottomrule
    \end{tabular}
    \caption{Examples of questions classified by their Subjectivity.}
    \label{tab:subjectivity_examples}
\end{table*}

\clearpage

\subsection{Subjectivity}
\cref{tab:subjectivity_examples} shows more examples classified by subjectivity. 

\section{Annotators' instructions}
We report here the instructions given to annotators when manually labeling the data. 

The annotators were given the exact same prompts as GPT. They were given the first iteration of the prompt engineering since the iterations were done by computing accuracy with respect to annotators' labels. Annotators were advised about the sources of the data considering that they could imply the potential presence of toxic / NSFW examples.  

\subsection{Annotation Instructions for Causality / Domain, Subjectivity / Cognitive Complexity / Needs / Open-Endedness}
\label{app:annotator_instructions}

DISCLAIMER - Sensitive / Offensive content
The questions were sourced from social media platforms, search engines, and conversational AI. This means that questions might contain offensive or sensitive content. By proceeding with the annotation task you are acknowledging this possibility and that you are comfortable reading and classifying the questions.

Task Description
In the file \{FILE\_NAME\}.xlsx there are \{N\} questions that were asked either on Quora, CharGPT, Bing, or Google.
The task is to classify each question depending on its \{FEATURE\}. The categories are explained below, as they were explained to GPT. 
The reason we are doing this is to understand whether the classification that we made via GPT is reliable / humans would agree with them. 

\{PROMPTS\}

\subsection{Annotation Instructions for Answer Grading}
\label{sec:iso_prompt}
In file \{FILENAME\} there are pairs of questions and answers. the questions were asked either on Quora, CharGPT, Bing, or Google. Judge the answers given the following rubric:

\begin{itemize}
    \item Effectiveness: Does the answer complete the goal specific to this user? How far is it with respect to the ideal perfect answer? 0: useless, 5: perfect
    \item Efficiency: Is the answer going to the point, without useless waste of time of the user? 0: a lot of useless information provided, 5: the answer provides just the right amount of information
    \item Satisfaction: Is the answer leaving the user satisfied, being empathetic, friendly, and pleasant to read? 0: annoying, 1: very pleasant

\end{itemize}

BE AWARE: because of the sources of the data, the prompts can contain toxic, disturbing, or NSFW topics. If you do not feel comfortable with this, feel free to skip the question or stop the annotation.

\end{document}